\documentclass[11pt,a4paper]{article}
\usepackage[hyperref]{emnlp2020}
\usepackage{times}
\usepackage{latexsym}

\usepackage{microtype}
\usepackage{graphicx}
\usepackage{comment}
\usepackage{amsmath,amssymb} 
\usepackage{color}
\usepackage{tablefootnote}

\usepackage{amsmath,amsfonts,bm}

\def\eqref#1{equation~\ref{#1}}

\def\1{\bm{1}}

\DeclareMathAlphabet{\mathsfit}{\encodingdefault}{\sfdefault}{m}{sl}
\SetMathAlphabet{\mathsfit}{bold}{\encodingdefault}{\sfdefault}{bx}{n}

\newcommand{\R}{\mathbb{R}}

\usepackage{url}
\usepackage{makecell}
\usepackage{graphicx}  
\usepackage{multirow}
\usepackage{tabularx}
\usepackage{booktabs}
\usepackage{colortbl}
\usepackage{arydshln}
\usepackage{subcaption}
\usepackage{xcolor}

\definecolor{asparagus}{rgb}{0.53, 0.66, 0.42}

\newcommand{\specialcelll}[2][l]{%
  \begin{tabular}[#1]{@{}l@{}}#2\end{tabular}}

\newcommand{\wv}{{\mathbf w}}
\newcommand{\vv}{{\mathbf v}}

\makeatletter
\newcommand\footnoteref[1]{\protected@xdef\@thefnmark{\ref{#1}}\@footnotemark}
\makeatother

\usepackage{algorithm,algpseudocode,adjustbox}
\aclfinalcopy 
\def\aclpaperid{2853} 

\title{\textsc{Hero}: Hierarchical Encoder for Video+Language \\ Omni-representation Pre-training}

\author{Linjie Li\thanks{\,\, Equal contribution.}, \hspace{0.03in} Yen-Chun Chen\footnotemark[1], \hspace{0.03in} Yu Cheng, \hspace{0.03in} Zhe Gan, \hspace{0.03in} Licheng Yu, \hspace{0.03in} Jingjing Liu\\
Microsoft Dynamics 365 AI Research\\
\small\tt{\{lindsey.li, yen-chun.chen, yu.cheng, zhe.gan, licheng.yu, jingjl\}@microsoft.com}}

\date{}

\begin{document}
\maketitle
\begin{abstract}
We present \textsc{Hero}, a novel framework for large-scale video+language omni-representation learning. \textsc{Hero} encodes multimodal inputs in a hierarchical structure, where \emph{local} context of a video frame is captured by a Cross-modal Transformer via multimodal fusion, and \emph{global} video context is captured by a Temporal Transformer. In addition to standard Masked Language Modeling (MLM) and Masked Frame Modeling (MFM) objectives, we design two new pre-training tasks: ($i$) Video-Subtitle Matching (VSM), where the model predicts both global and local temporal alignment; and ($ii$) Frame Order Modeling (FOM), where the model predicts the right order of shuffled video frames. \textsc{Hero} is jointly trained on HowTo100M and large-scale TV datasets to gain deep understanding of complex social dynamics with multi-character interactions. Comprehensive experiments demonstrate that \textsc{Hero} achieves new state of the art on multiple benchmarks over Text-based Video/Video-moment Retrieval, Video Question Answering (QA), Video-and-language Inference and Video Captioning tasks across different domains. We also introduce two new challenging benchmarks How2QA and How2R for Video QA and Retrieval, collected from diverse video content over multimodalities.\footnote{Code and new datasets will be released at \url{https://github.com/linjieli222/HERO}.}
\end{abstract}

\section{Introduction}
\label{sec:intro}
Inspired by BERT~\cite{devlin2018bert}, large-scale multimodal pre-training has prevailed in the realm of vision-and-language research~\cite{lu2019vilbert,tan2019lxmert,chen2019uniter}.
There are many early players in the area, including ViLBERT~\citep{lu2019vilbert}, LXMERT~\citep{tan2019lxmert}, UNITER~\citep{chen2019uniter}, VL-BERT~\cite{su2019vl} and Unicoder-VL~\cite{li2019unicoder}. However, most large-scale pre-trained models are tailored for static images, not dynamic videos. 
VideoBERT~\cite{sun2019videobert} is the first to apply BERT to learn joint embedding for video-text pairs. But since only discrete tokens are used to represent video frames, rich video frame features are not fully utilized. To remedy this, CBT~\cite{sun2019contrastive} proposes to use a contrastive loss, but mainly for video representation learning alone, with text input only considered as side information. UniViLM~\cite{luo2020univilm} takes a step further and considers both understanding and generation tasks.

Several constraints inherently limit the success of existing models. ($i$) Most model designs are direct adaptation of BERT, taking simple concatenation of subtitle sentences and visual frames as input, while losing the temporal alignment between video and text modalities. 
($ii$) Pre-training tasks are directly borrowed from image+text pre-training methods, without exploiting the sequential nature of videos.
($iii$) Compared to diverse image domains investigated in existing work, video datasets used in current models are restricted to cooking or narrated instructional videos~\cite{miech2019howto100m}, excluding video sources that contain dynamic scenes and complex social interactions.

\begin{figure*}[!t]
\centering
  \includegraphics[width=1.0\linewidth]{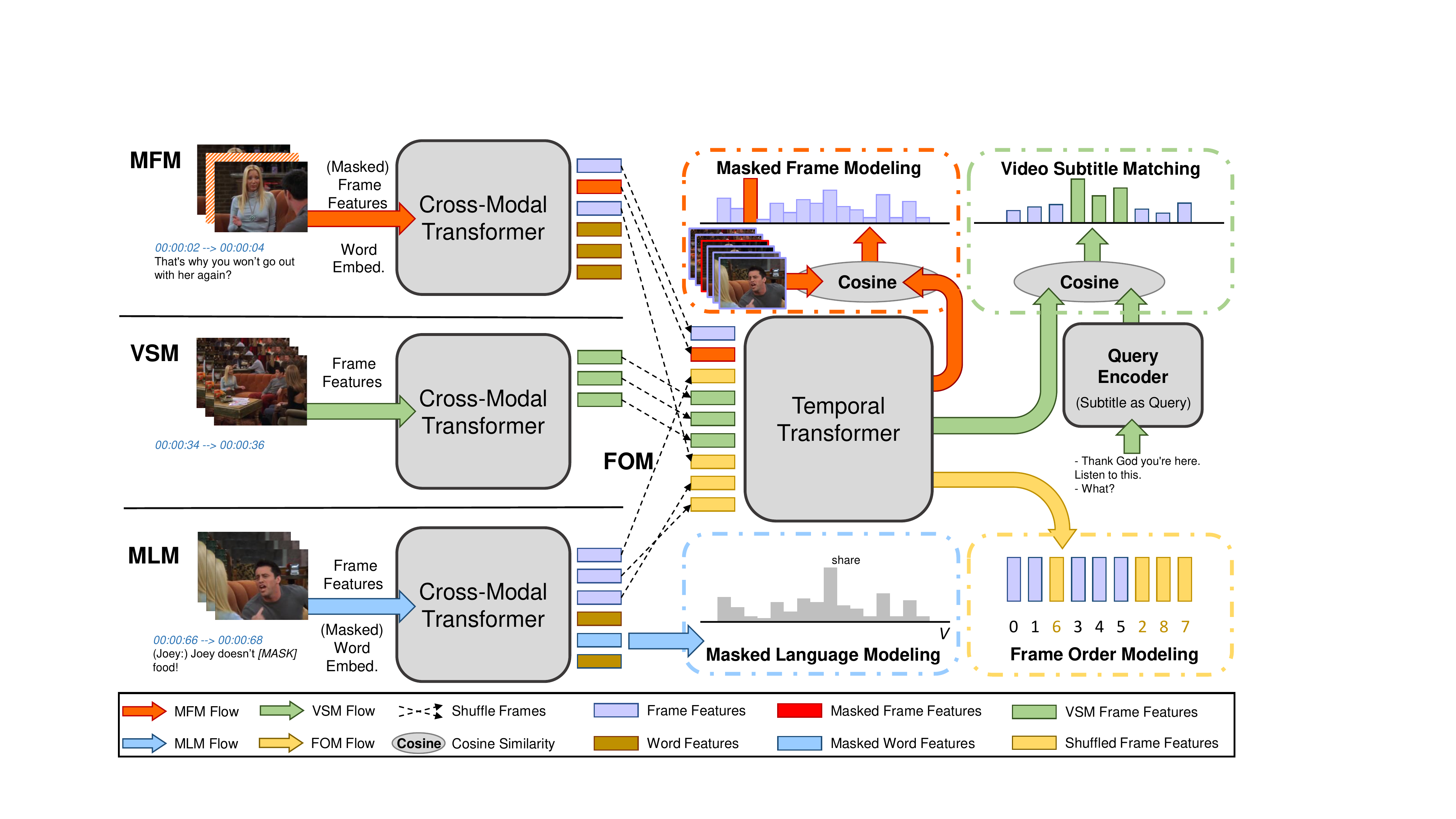}
 \caption{\textsc{Hero} Architecture (best viewed in color), consisting of Cross-Modal Transformer and Temporal Transformer, learned via
four pre-training tasks hierarchically. 
Initial frame features are obtained by SlowFast and ResNet feature extractors, and word embeddings are learned via an embedding layer initialized from RoBERTa.} 
  \label{fig:model}
\end{figure*}

To tackle these challenges, we present a new video-and-language large-scale pre-training framework - \textsc{Hero} (\textbf{H}ierarchical \textbf{E}ncode\textbf{R} for \textbf{O}mni-representation learning). As illustrated in Figure~\ref{fig:model}, \textsc{Hero} takes as input a sequence of video clip frames and their accompanying subtitle sentences.\footnote{ASR can be applied when subtitles are unavailable.} Instead of adopting a flat BERT-like encoder, \textsc{Hero} encodes multimodal inputs in a hierarchical fashion, with ($i$) a \emph{Cross-modal} Transformer to fuse a subtitle sentence and its accompanying local video frames, followed by ($ii$) a \emph{Temporal} Transformer to obtain a sequentially contextualized embedding for each video frame, using all the surrounding frames as global context. The proposed hierarchical model first absorbs visual and textual local context on frame level, which is then transferred to a global video-level temporal context. Experiments show that this novel model design achieves better performance than a flat BERT-like architecture. 

Four pre-training tasks are designed for \textsc{Hero}: ($i$) Masked Language Modeling (MLM); ($ii$) Masked Frame Modeling (MFM); ($iii$) Video-Subtitle Matching (VSM); and ($iv$) Frame Order Modeling (FOM). Compared to prior work, the key novelty is VSM and FOM, which encourage explicit temporal alignment between multimodalities as well as full-scale exploitation of the sequential nature of video input. 
In VSM, the model considers not only global alignment (predicting whether a subtitle matches the input video clip), but also local temporal alignment (retrieving the moment where the subtitle should be localized in the video clip). In FOM, we randomly select and shuffle a subset of video frames, and the model is trained to restore their original order. Extensive ablation studies demonstrate that both VSM and FOM play a critical role in video+language pre-training. 

To empower the model with richer knowledge beyond instructional videos used in prior work, we jointly train \textsc{Hero} with both HowTo100M (narrated instructional videos)~\cite{miech2019howto100m} and a large-scale TV dataset (containing TV episodes spanning across different genres)~\cite{lei2018tvqa,lei2019tvqaplus,lei2020tvr,liu2020violin}. Compared to factual descriptions in HowTo100M, the TV dataset contains more complex plots that require comprehensive interpretation of human emotions, social dynamics and causal relations of events, making it a valuable supplement to HowTo100M and a closer approximation to real-life scenarios. 

Existing pre-trained models are evaluated on YouCook2~\citep{zhou2018towards} and MSR-VTT~\citep{xu2016msr} datasets. YouCook2 focuses on cooking videos only, and the captions in MSR-VTT are very simple. To evaluate our model on more challenging benchmarks, we collect two new datasets on video-moment retrieval and question answering, \emph{How2R} and \emph{How2QA}.
In addition, we evaluate \textsc{Hero} on popular retrieval and QA tasks such as TVR~\citep{lei2020tvr} and TVQA~\citep{lei2018tvqa}, where \textsc{Hero} outperforms existing models by a large margin. We further demonstrate the generalizability of our model by adapting it to ($i$) diverse downstream tasks: video-and-language inference and video captioning tasks, achieving new state of the art on VIOLIN~\citep{liu2020violin} and  TVC~\citep{lei2020tvr} benchmarks;  ($ii$) different video types: single-channel videos (video-only) and  multi-channel videos (video + subtitle), reporting superior performance over existing state of the art on DiDeMo ~\citep{didemo} and MSR-VTT. 

Our main contributions are summarized as follows. ($i$) We present \textsc{Hero}, a hierarchical Transformer-based model for video+language representation learning. ($ii$) We propose new pre-training tasks VSM and FOM, which complement MLM and MRM objectives by better capturing temporal alignment between multimodalities in both global and local contexts. ($iii$) Different from previous work that mainly relies on HowTo100M, we include additional video datasets for pre-training, encouraging the model to learn from richer and more divserse visual content. ($iv$) We collect two new datasets based on HowTo100M for video-moment retrieval/QA, and will release the new benchmarks to foster future study. \textsc{Hero} achieves new state of the art across all the evaluated tasks. 

\section{Related Work}

\label{sec:related}

Since the birth of BERT~\cite{devlin2018bert}, there has been continuing advancement in language model pre-training, such as XLNet~\cite{yang2019xlnet}, RoBERTa~\cite{liu2019roberta}, ALBERT~\cite{lan2019albert},
UniLM~\cite{dong2019unified}, and T5~\cite{raffel2019exploring}, which epitomizes the superb power of large-scale pre-training. Satellited around BERT, there is parallel growing interest in model compression~\cite{sun2019patient, shen2019q} and extension to generation tasks~\cite{chen2019distilling, wang2019bert}.

Branching out from language processing to multimodal, subsequent studies also emerge in vision+language space. 
Prominent work includes ViLBERT~\cite{lu2019vilbert}, LXMERT~\cite{tan2019lxmert},
VL-BERT~\cite{su2019vl}, Unicoder-VL~\cite{li2019unicoder}, B2T2~\cite{alberti2019fusion}, UNITER~\cite{chen2019uniter} and VILLA~\cite{gan2020large}. A detailed review can be found in Appendix \ref{app:vl_overview}. 

Contrast to the boom in image+text area, pre-training for video+language is still in its infancy. So far, VideoBERT~\cite{sun2019videobert}, CBT~\cite{sun2019contrastive}, MIL-NCE~\cite{miech2020end}, ActBERT~\cite{zhu2020actbert} and UniViLM~\cite{luo2020univilm} are the only existing work exploring this space, covering downstream tasks from text-based video retrieval~\citep{zhou2018towards, xu2016msr-vtt} and video question answering~\citep{maharaj2017dataset,lei2019tvqaplus} to video captioning~\cite{zhou2018end}.

In this paper, we aim to propel video+language omni-representation learning in four dimensions: ($i$) better model architecture design; ($ii$) better pre-training task design; ($iii$) diversification of training corpora; and ($iv$) new high-quality benchmarks for downstream evaluation.

\section{Hierarchical Video+Language Encoder}
\label{sec:method}

In this section, we explain the proposed \textsc{Hero} architecture and the four pre-training tasks in detail.

\subsection{Model Architecture}\label{sec:method_overview}

Model architecture of \textsc{Hero} is illustrated in Figure~\ref{fig:model}, which takes the frames of a video clip and the textual tokens of subtitle sentences as inputs.
They are fed into a Video Embedder and a Text Embedder to extract initial representations. \textsc{Hero} computes contextualized video embeddings in a hierarchical procedure. 
First, \emph{local} textual context of each visual frame is captured by a Cross-modal Transformer, computing the contextualized multi-modal embeddings between a subtitle sentence and its associated visual frames. 
The encoded frame embeddings of the whole video clip are then fed into Temporal Transformer to learn the \emph{global} video context and obtain the final contextualized video embeddings.

\paragraph{Input Embedder}
We denote visual frames of a video clip as $\mathbf{v} = \{v_i\}^{N_v}_{i=1}$ and its subtitle as $\mathbf{s} = \{s_i\}^{N_s}_{i=1}$ ($N_v$ is the number of visual frames in a video clip and $N_s$ is the number of sentences in each subtitle). For \textit{Text Embedder}, we follow \citet{liu2019roberta} and tokenize a subtitle sentence $s_i$ into a sequence of WordPieces \citep{wu2016google}, \emph{i.e.}, $\mathbf{w}_{s_i} = \{w_{s_i}^{j}\}_{j=1}^L$ ($L$ is the number of tokens in $s_i$). The final representation for each sub-word token is obtained via summing up its token embedding and position embedding, followed by a layer normalization (LN) layer. For \textit{Video Embedder}, we first use ResNet~\citep{he2016deep} pre-trained on ImageNet~\citep{deng2009imagenet} and SlowFast~\citep{feichtenhofer2019slowfast} pre-trained on Kinetics~\citep{kay2017kinetics} to extract 2D and 3D visual features for each video frame. These features are concatenated as visual features and fed through a fully-connected (FC) layer to be projected into the same lower-dimensional space as token embeddings. Since video frames are sequential, their position embeddings can be calculated in the same way as in the Text Embedder. The final embedding of a frame is obtained by summing up FC outputs and position embeddings and then passing through an LN layer. After Input Embedder, token and frame embeddings for $\mathbf{w}_{s_i}$ and $\mathbf{v}_{s_i}$\footnote{$\mathbf{v}_{s_i} = \{v_{s_i}^j\}_{j=1}^K$ denotes the set of visual frames paired with subtitle sentence $s_i$, based on their timestamps. Refer to Appendix
~\ref{app: pairing} 
for details. 

} are denoted as $\mathbf{W}_{s_i}^{emb} \in \R^{L\times d}$ and $\mathbf{V}_{s_i}^{emb} \in \R^{K \times d}$ ($d$ is the hidden size).

\paragraph{Cross-modal Transformer}
To utilize the inherent alignment between subtitles and video frames, for each subtitle sentence $s_i$, we first learn contextualized embeddings between the corresponding tokens $\mathbf{w}_{s_i}$ and its associated visual frames $\mathbf{v}_{s_i}$ through cross-modal attention. Inspired by the recent success~\citep{chen2019uniter,lu2019vilbert} of using Transformer~\citep{vaswani2017attention} for multimodal fusion, we also use a multi-layer Transformer here. The outputs from Cross-modal Transformer is a sequence of contextualized embeddings for each subtitle token and each video frame:
\begin{align}
    \mathbf{V}_{s_i}^{cross}, \mathbf{W}_{s_i}^{cross} = f_{cross}(\mathbf{V}_{s_i}^{emb}, \mathbf{W}_{s_i}^{emb})\,,
\end{align}
where $f_{cross}(\cdot,\cdot)$ denotes the Cross-modal Transformer, $\mathbf{V}_{s_i}^{cross} \in \R^{K \times d}$ and $\mathbf{W}_{s_i}^{cross} \in \R^{L\times d}$. 

\paragraph{Temporal Transformer}
After collecting all the visual frame embeddings $\mathbf{V}^{cross} = \{\mathbf{V}_{s_i}^{cross}\}_{i=1}^{N_s} \in \R^{N_v \times d}$ from the output of Cross-modal Transformer, we use another Transformer as temporal attention to learn contextualized video embeddings from the global context of a video clip. To avoid losing positional information, we use residual connection~\citep{he2016deep} to add back $\mathbf{V}^{emb}\in \R^{N_v\times d}$. The final contextualized video embeddings are calculated as:
\begin{align}
    \mathbf{V}^{temp} = f_{temp}(\mathbf{V}^{emb} + \mathbf{V}^{cross})\,,
\end{align}
where $f_{temp}(\cdot)$ denotes the Temporal Transformer, and $\mathbf{V}^{temp} \in \R^{N_v \times d}$. Compared to flat BERT-like encoder, which directly concatenates all textual tokens and visual frames as inputs, the proposed model effectively utilizes the temporal alignment between subtitle sentences and video frames for multimodal fusion in a more fine-grained manner. In the experiments, we show that our model design far outperforms a flat BERT-like baseline. 

\subsection{Pre-training Tasks}\label{sec:pretraining_tasks}
We introduce four tasks for pre-training. During training, we sample one task per mini-batch to prevent different tasks from corrupting each others' input. As shown in Figure~\ref{fig:model}, MFM and MLM are in analogy to BERT~\citep{devlin2018bert}.
Word masking is realized by replacing a word with special token \texttt{[MASK]}, and frame masking by replacing a frame feature vector with zeros.
Following \citet{chen2019uniter}, we only mask one modality each time while keeping the other modality intact.
VSM is designed to learn both \emph{local} alignment (between visual frames and a subtitle sentence) and \emph{global} alignment (between a video clip and a sequence of subtitle sentences). 
FOM is designed to model sequential characteristics of video, by learning the original order of randomly reordered frames. 

\vspace{5pt}
\subsubsection{Masked Language Modeling}\label{sec:mlm}
The inputs for MLM include: ($i$) sub-word tokens from the $i$-th subtitle sentence $\wv_{s_i}$; ($ii$) visual frames $\vv_{s_i}$ aligned with $\wv_{s_i}$; and ($iii$) mask indices $\mathbf{m} \in \mathbb{N}^M$.\footnote{$\mathbb{N}$ is a natural number, $M$ is the number of masked tokens, and $\mathbf{m}$ is the set of masked indices.}

In MLM, we randomly mask out input words with a probability of 15\%, and replace the masked tokens $\mathbf{w}_{s_i}^{\mathbf{m}}$ with special tokens \texttt{[MASK]}.\footnote{Following BERT, we decompose the 15\% randomly masked-out words into 10\% random words, 10\% unchanged, and 80\% \texttt{[MASK]}.}
The goal is to predict these masked words based on the observation of their surrounding words $\mathbf{w}_{s_i}^{\setminus \mathbf{m}}$ and the visual frames aligned with the sentence $\mathbf{v}_{s_i}$, by minimizing the negative log-likelihood:
\begin{equation}
    \mathcal{L}_{\text{MLM}}(\theta) = -\mathbb{E}_{D} \log P_{\theta}(\mathbf{w}_{s_i}^{\mathbf{m}} | \mathbf{w}_{s_i}^{\setminus \mathbf{m}}, \mathbf{v}_{s_i})\,,
\end{equation}
where $\theta$ denotes trainable parameters. 
Each pair $(\mathbf{w}_{s_i}, \mathbf{v}_{s_i})$ is sampled from the training set $D$.

\vspace{5pt}
\subsubsection{Masked Frame Modeling} 
Similar to MLM, we also sample frames and mask their visual features with a probability of 15\%. However, the difference is that MLM is performed on a local context (\emph{i.e.}, the output of Cross-modal Transformer), while MFM is performed on a global context (\emph{i.e.}, the output of Temporal Transformer). 
The model is trained to reconstruct masked frames $\mathbf{v}_{\mathbf{m}}$, given the remaining frames $\mathbf{v}_{\setminus \mathbf{m}}$ and all the subtitle sentences $\mathbf{s}$. 
The visual features of masked frames are replaced by zeros.
Unlike textual tokens that are represented as discrete labels, visual features are high-dimensional and continuous, thus cannot be supervised via class likelihood.
Instead, we propose two variants for MFM, which share the same objective base:
\begin{equation}
    \mathcal{L}_{\text{MFM}}(\theta) = \mathbb{E}_{D} f_{\theta}(\mathbf{v}_{\mathbf{m}} | \mathbf{v}_{\setminus \mathbf{m}}, \mathbf{s})\,.
\end{equation}

\paragraph{Masked Frame Feature Regression (MFFR)}
MFFR learns to regress the output on each masked frame $\mathbf{v}_\mathbf{m}^{(i)}$ to its visual features. Specifically, we apply an FC layer to convert the output frame representations into a vector $h_{\theta}(\mathbf{v}_\mathbf{m}^{(i)})$ of the same dimension as the input visual feature $r(\mathbf{v}_\mathbf{m}^{(i)})$.
Then we apply L2 regression between the two: $f_{\theta}(\mathbf{v}_\mathbf{m} | \mathbf{v}_{\setminus \mathbf{m}}, \mathbf{s}) = \sum_{i=1}^M \| h_{\theta}(\mathbf{v}_\mathbf{m}^{(i)}) - r(\mathbf{v}_\mathbf{m}^{(i)}) \|_2^2$. 

\vspace{5pt}
\paragraph{Masked Frame Modeling with Noise Contrastive Estimation (MNCE)}
Instead of directly regressing the real values of masked visual features, we use the softmax version of Noise Contrastive Estimation (NCE) loss~\cite{jozefowicz2016exploring}, which is widely adopted in self-supervised representation learning~\citep{sun2019contrastive,hjelm2018learning,oord2018representation}. NCE loss encourages the model to identify the correct frame (given the context) compared to a set of negative distractors.

Similar to MFFR, we feed the output of the masked frames $\mathbf{v}_\mathbf{m}^{(i)}$ into an FC layer to project them into a vector $g_{\theta}(\mathbf{v}_\mathbf{m}^{(i)})$. Moreover, we randomly sample frames from the output of unmasked frames as negative distractors $\mathbf{v}_\mathbf{neg} = \{\mathbf{v}_\mathbf{neg}^{(j)} | \mathbf{v}_\mathbf{neg}^{(j)} \in  \mathbf{v}_{\setminus \mathbf{m}}\}$, which are also transformed through the same FC layer as ${g_{\theta}(\mathbf{v}_\mathbf{neg}^{(j)})}$. The final objective minimizes the NCE loss: 
$f_{\theta}(\mathbf{v}_\mathbf{m} | \mathbf{v}_{\setminus \mathbf{m}}, \mathbf{s}) = \sum_{i=1}^M \log \text{NCE}(g_{\theta}(\mathbf{v}_\mathbf{m}^{(i)}) | g_{\theta}(\mathbf{v}_\mathbf{neg}) )$.

\vspace{5pt}
\subsubsection{Video-Subtitle Matching} 
The inputs to VSM are: $(i)$ a sampled query $s_q$ from all subtitle sentences; $(ii)$ the whole video clip $\mathbf{v}$; and $(iii)$ the remaining subtitle sentences $\mathbf{s}_{\setminus q}$ for the video clip. We expect the model to learn: $(i)$ \emph{local alignment} - the start and end index $y_{st}, y_{ed} \in \{1,..., N_v\}$, indicating the span of visual frames aligned with the query;\footnote{
Timestamps are used to perform local alignment, which are either included with video (e.g., TV) or generated by ASR (e.g., HowTo100M). Refer to Appendix \ref{app: pairing} for details.} and $(ii)$ \emph{global alignment} - to which video the sampled query is matched.

In VSM, we follow XML~\citep{lei2020tvr} to compute the matching scores between the query and visual frames at both local and global levels. Specifically, we extract the output of Temporal Transformer as the final visual frame representation $\mathbf{V}^{temp}\in \R^{N_v\times d}$. The query is fed into Cross-modal Transformer to compute its textual representations $\mathbf{W}^{cross}_{s_q} = f_{cross} (\mathbf{0}, \mathbf{W}^{embed}_{s_q})$. Based on this, we use a query encoder \citep{lei2020tvr}, consisting of a self-attention layer, two linear layers and an LN layer, to obtain the final query vector $\mathbf{q}\in \R^d$ from $\mathbf{W}^{cross}_{s_q}$.

\paragraph{Local Alignment}
The local query-video matching score is computed using dot product: \begin{equation}
    S_{local}(s_q, \vv) = \mathbf{V}^{temp}\mathbf{q}\in \R^{N_v}\,.
\end{equation}
Two trainable 1D convolution filters are applied to the scores, followed by a softmax layer, to generate two probability vectors $\mathbf{p}_{st}, \mathbf{p}_{ed} \in \mathbb{R}^{N_v}$, representing the probabilities of every position being the start and
end of the ground-truth span.
During training, we sample 15\% subtitle sentences as queries for each video, and use the cross-entropy loss to predict the start and end index for local alignment: 
\begin{align*}
\mathcal{L}_{local} &= -\mathbb{E}_{D}\log(\mathbf{p}_{st}[y_{st}]) + \log(\mathbf{p}_{ed}[y_{ed}])\,,
\end{align*}
where $\mathbf{p}[y]$ denotes indexing the $y$-th element of the vector $\mathbf{p}$.

Note that, XML computes the query-video matching score for each modality separately, and the final matching score is the sum of the two scores. In our \textsc{Hero} model, multimodal fusion is performed in a much earlier stage.

\paragraph{Global Alignment}
The global matching score is computed by max-pooling the cosine similarities between each frame and the query:
\begin{equation}
 S_{global}(s_q, \vv) = \max\left(\frac{\mathbf{V}^{temp}}{||\mathbf{V}^{temp}||}\frac{\mathbf{q}}{||\mathbf{q}||}\right)\,. \label{eqn:global_alignment}
\end{equation}
We use a combined hinge loss $\mathcal{L}_h$~\citep{yu2018mattnet} over positive and negative query-video pairs. For each positive pair $(s_q, \vv)$, we replace $\vv$ or $s_q$ with one other sample from in the same mini-batch to construct two sets of negative examples: $(s_q, \hat{\vv})$ and $(\hat{s}_q, \vv)$. The training loss is specified as:
\begin{align}
&\mathcal{L}_{h} (S_{pos}, S_{neg}) = \max(0, \delta + S_{neg}- S_{pos})\,, \nonumber \\
    &\mathcal{L}_{global} = -\mathbb{E}_{D}[\mathcal{L}_{h}( S_{global}(s_q, \vv), S_{global}(\hat{s}_q, \vv)) \nonumber\\ 
    &+ \mathcal{L}_{h}( S_{global}(s_q, \vv), S_{global}(s_q, \hat{\vv}))]\,,
\end{align}
where $\delta$ is the margin hyper-parameter. 
The final loss $\mathcal{L}_{\text{VSM}} = \lambda_1 \mathcal{L}_{local} + \lambda_2 \mathcal{L}_{global}$, where $\lambda_1$ and $\lambda_2$ are hyper-parameters balancing the two terms. 

\vspace{5pt}
\subsubsection{Frame Order Modeling} 
The inputs for FOM are: $(i)$ all subtitle sentences $\mathbf{s}$; $(ii)$ visual frames $\vv$; and $(iii)$ the reorder indices $\mathbf{r}=\{r_i\}_{i=1}^R\in \mathbb{N}^R$.\footnote{$R$ is the number of reordered frames, and $\mathbf{r}$ is the set of reorder indices.} We randomly select 15\% of the frames to be shuffled, and the goal is to reconstruct their original timestamps, denoted as $\mathbf{t} = \{t_i\}_{i = 1}^{R}$, where $t_i \in \{1, ..., N_v\}$. We formulate FOM as a classification problem, where $\mathbf{t}$ is the ground-truth labels of the reordered frames. 

Specifically, reordering happens after the multimodal fusion of subtitle and visual frames.
The reordered features are fed into Temporal Transformer to produce reordered visual frame embeddings $\mathbf{V}^{temp}_{\mathbf{r}}$. 
These embeddings are transformed through an FC layer, followed by a softmax layer to produce a probability matrix $\mathbf{P} \in \mathbb{R}^{N_v \times N_v}$, where each column $\mathbf{p}_{i} \in \mathbb{R}^{N_v}$ represents the scores of $N_v$ timestamp classes that the $i$-th timestamp belongs to. The final objective is to minimize the the negative log-likelihood (cross-entropy loss):
\begin{equation}
    \mathcal{L}_{\text{FOM}} = -\mathbb{E}_{D} \textstyle{\sum_{i=1}^R} \log \mathbf{P}[r_i, t_i]\,.
\end{equation}

\section{Experiments}
\label{exps}
 In this section, we describe comprehensive experiments on downstream tasks and provide ablation studies for in-depth analysis of different pre-training settings.

To validate the effectiveness of \textsc{Hero}, we evaluate on a wide variety of downstream tasks, including Text-based Video/ Video-moment Retrieval, Video Question Answering, Video-and-language Inference, and Video Captioning. We consider 6 existing benchmarks: TVR~\citep{lei2020tvr}, TVQA~\citep{lei2018tvqa}, VIOLIN~\citep{liu2020violin}, TVC~\citep{lei2020tvr}, DiDeMo~\citep{didemo}, and MSR-VTT~\citep{xu2016msr-vtt}. Detailed descriptions and evaluation metrics on each task can be found in Appendix~\ref{app:downstream_data}.

\subsection{Pre-training Datasets}
Our pre-training dataset is composed of 7.6M video clips with their accompanying subtitles from TV and HowTo100M datasets. We exclude all the videos that appear in the downstream tasks to avoid contamination in evaluation. 

\paragraph{TV Dataset}\citep{lei2018tvqa} was built on 6 popular TV shows across 3 genres: medical dramas, sitcoms and crime shows. It contains 21,793 video clips from 925 episodes. Each video clip is 60-90 seconds long, covering long-range scenes with complex character interactions and social/professional activities. Dialogue for each video clip is also provided.

\paragraph{HowTo100M Dataset}\citep{miech2019howto100m} was collected from YouTube, mostly instructional videos. It contains 1.22 million videos, with activities falling into 12 categories (e.g., Food \& Entertaining, Home \& Garden, Hobbies \& Crafts). Each video is associated with a narration as subtitles that are either written manually or from an Automatic Speech Recognition (ASR) system. The average duration of videos in HowTo100M is 6.5 minutes. We cut the videos into 60-second clips to make them consistent with the TV dataset, and exclude videos in non-English languages. These pre-processing steps result in a subset of 7.56M video clips, accompanied with English subtitles.

\begin{table*}[t!]
\centering
\resizebox{.99\textwidth}{!}{
\begin{tabular}{l p{0.02\textwidth} l c  c  c c c c c c}
\hline
Pre-training Data & & Pre-training Tasks&  \multicolumn{3}{c}{TVR} & TVQA & \multicolumn{3}{c}{How2R} & How2QA\\ 
 \cmidrule(lr){4-6} \cmidrule(lr){7-7} \cmidrule(lr){8-10}  \cmidrule(lr){11-11} 
& & & R@1  & R@10 & R@100  &  Acc. & R@1  & R@10 & R@100 & Acc. \\ 
\hline
\multirow{5}{*}{TV} &  \small{1} & MLM & 2.92  & 10.66  &17.52  & 71.25 & 2.06 & 9.08 & 14.45 & 69.79 \\
& \small{2} & MLM + MNCE & 3.13 & 10.92 & 17.52 & 71.99 & 2.15 & 9.27 & 14.98 & 70.13 \\
& \small{3} & MLM + MNCE + FOM & 3.09 & 10.27 & 17.43 & \cellcolor[gray]{.8}72.54 & 2.36 & 9.85 & 15.97 & 70.85 \\
& \small{4} & MLM + MNCE + FOM + VSM & \cellcolor[gray]{.6}4.44 & \cellcolor[gray]{.6}14.69 & \cellcolor[gray]{.6}22.82  & \cellcolor[gray]{.6}72.75  & \cellcolor[gray]{.6}2.78 & \cellcolor[gray]{.6}10.41 & \cellcolor[gray]{.6}18.77 & \cellcolor[gray]{.6} 71.36\\
& \small{5} & MLM + MNCE + FOM + VSM + MFFR &  \cellcolor[gray]{.6}4.44& \cellcolor[gray]{.8}14.29 & \cellcolor[gray]{.8}22.37 & \cellcolor[gray]{.6}72.75 & \cellcolor[gray]{.8}2.73 &\cellcolor[gray]{.8}10.12 & \cellcolor[gray]{.8}18.05 & \cellcolor[gray]{.6} 71.36\\
\hline
Howto100M & \small{6} & MLM + MNCE + FOM + VSM & 3.81& 13.23 & 21.63 & 73.34 & 3.54 & \textbf{12.90} & 20.85 &  73.68 \\
\hline
TV + HowTo100M & \small{7} & MLM + MNCE + FOM + VSM & \textbf{5.13} &  \textbf{16.26} & \textbf{24.55} &   \textbf{74.80} & \textbf{3.85} & 12.73 & \textbf{21.06}  & \textbf{73.81} \\
\hline
\end{tabular}
}
\caption{Evaluation on pre-training tasks and datasets. Dark and light grey colors highlight the top and second best results across all the tasks trained with TV Dataset. The best results are in bold.}
\label{table:pretrain_ablation}
\vspace{-2mm}
\end{table*}

\subsection{New Benchmarks}
Existing benchmarks are mostly built on videos from either a single domain or a single modality. In order to evaluate on diverse video content that reflects multimodality challenges, we introduce two new datasets as additional benchmarks: \emph{How2R} for text-based video-moment retrieval, and \emph{How2QA} for video question answering.

\paragraph{How2R}
Amazon Mechanical Turk (AMT) is used to collect annotations on HowTo100M videos. Figure 6a in Appendix 
shows the interface for annotation. We randomly sample 30k 60-second clips from 9,421 videos and present each clip to the turkers, who are asked to select a video segment containing a single, self-contained scene. After this segment selection step, another group of workers are asked to write descriptions for each displayed segment. Narrations are not provided to the workers to ensure that their written queries are based on visual content only.  These final video segments are 10-20 seconds long on average, and the length of queries ranges from 8 to 20 words.
 
From this process, we have collected 51,390 queries for 24k 60-second clips from 9,371 videos in HowTo100M, on average 2-3 queries per clip. We split the video clips and its associated queries into 80\% train, 10\% val and 10\% test. 

\paragraph{How2QA}
To collect another dataset for video QA task, we present the same set of selected video clips to another group of AMT workers for multi-choice QA annotation. Each worker is assigned with one video segment and asked to write one question with four answer candidates (one correct and three distractors). Similarly, narrations are hidden from the workers to ensure the collected QA pairs are not biased by subtitles. 

We observe that human-written negative answers suffer from serious bias (\emph{i.e.}, models can learn to predict correctly without absorbing any information from the video or subtitles). To mitigate this, we use adversarial matching~\citep{zellers2019recognition} to replace one of the three written negative answers by a correct answer from another question that is most relevant to the current one. 
Similar to TVQA, we also provide the start and end points for the relevant moment for each question. After filtering low-quality annotations, the final dataset contains 44,007 QA pairs for 22k 60-second clips selected from 9035 videos.
We split the data into 80\% train, 10\% val and 10\% test sets. More details about data collection can be found in Appendix \ref{app:data_analysis}.

\subsection{Ablation Study}
We analyze the effectiveness of model design, especially different combinations of pre-training tasks and datasets, through extensive ablation studies.

\begin{table*}[t!]
\renewcommand\thetable{3}
\begin{subtable}[h]{\textwidth}
\centering
\resizebox{1.0\textwidth}{!}{
\begin{tabular}{l c c  c  c c c c c c c c c c}
\hline
Method \textbackslash Task & \multicolumn{3}{c}{TVR} &   \multicolumn{3}{c}{How2R} & TVQA & How2QA & VIOLIN & \multicolumn{4}{c}{TVC} \\ 
 \cmidrule(lr){2-4} \cmidrule(lr){5-7} \cmidrule(lr){8-8} \cmidrule(lr){9-9} \cmidrule(lr){10-10} \cmidrule(lr){11-14} 
& R@1 & R@10 & R@100  &  R@1 & R@10 & R@100 & Acc. &  Acc. &  Acc. & Bleu & Rouge-L & Meteor & Cider\\
\hline
SOTA Baseline &3.25 & 13.41 & 30.52  & 2.06 & 8.96 & 13.27 & 70.23 & - & 67.84  & 10.87 & 32.81 & 16.91 & 45.38 \\
\hline
\textsc{Hero} &  \textbf{6.21} &  \textbf{19.34} & \textbf{36.66} &  \textbf{3.85} & \textbf{12.73} & \textbf{21.06}  & \textbf{73.61} & \textbf{73.81} & \textbf{68.59} & \textbf{12.35} & \textbf{34.16} & \textbf{17.64} & \textbf{49.98}  \\
\hline
\end{tabular}
}
\caption{Results on multi-channel (video+subtitle) tasks: TVR\footnoteref{note:tvr_test}, How2R, TVQA, How2QA, VIOLIN and TVC.}
\label{tab:sota_a}
\end{subtable}

\vspace{15pt}

\begin{subtable}[h]{\textwidth}
\centering
\resizebox{0.85\textwidth}{!}{
\begin{tabular}{l c ccc  ccc ccc ccc}
\hline
Method \textbackslash Task & \multicolumn{3}{c}{DiDeMo} & \multicolumn{3}{c}{DiDeMo w/ ASR} &   \multicolumn{3}{c}{MSR-VTT} & \multicolumn{3}{c}{MSR-VTT w/ ASR}\\ 
 \cmidrule(lr){2-4} \cmidrule(lr){5-7} \cmidrule(lr){8-10} \cmidrule(lr){11-13}
& R@1 & R@10 & R@100  &  R@1 & R@10 & R@100 & R@1 & R@5 & R@10 & R@1 & R@5 & R@10\\
\hline
SOTA Baseline &1.59 & 6.71 & 25.44  & - & - & - &  14.90 &  40.20 & 52.80 &  - & - & - \\
\hline
\textsc{Hero} &  \textbf{2.14} & \textbf{11.43} & \textbf{36.09}  &  \textbf{3.01} & \textbf{14.87} & \textbf{47.26}  &  \textbf{16.80} & \textbf{43.40} & \textbf{57.70} & \textbf{20.50} & \textbf{47.60} & \textbf{60.90}\\
\hline
\end{tabular}
}
\caption{Results on DiDeMo and MSR-VTT with video-only inputs (single-channel), compared with ASR-augmented inputs (multi-channel).}
\label{tab:sota_b}
\end{subtable}
\caption{Results on the test set of six downstream tasks, compared to task-specific state-of-the-art (SOTA) models: XML~\citep{lei2020tvr} for TVR, How2R and DiDeMo, HowTo100M~\citep{miech2019howto100m} for MSR-VTT, STAGE~\citep{lei2019tvqaplus} for TVQA (inapplicable to How2QA due to region-level features), Multi-stream~\citep{liu2020violin} for VIOLIN, and MMT~\citep{lei2020tvr} for TVC.}
\label{table:results_tab}
\end{table*}

\paragraph{Optimal Setting of  Pre-training Tasks}
To search for the optimal setting of pre-training tasks, we conduct a series of extensive ablation studies to test each setting, using video-moment retrieval and QA downstream tasks as evaluation.
Table~\ref{table:pretrain_ablation} summarizes ablation results on TVR, TVQA, How2R and How2QA under different pre-training settings. Models are trained on TV dataset only for computational efficiency. Compared to using MLM only (L1 in Table~\ref{table:pretrain_ablation}), adding MNCE (L2) shows improvement on all downstream tasks. The best performance is achieved by MLM + MNCE + FOM + VSM (L4).

\paragraph{Effect of FOM and VSM}
When MLM, MNCE and FOM are jointly trained (L3), there is a large performance gain on TVQA, and significant improvement on How2R and How2QA. Comparable results are achieved on TVR. This indicates that FOM, which models sequential characteristics of video frames, can effectively benefit downstream tasks that rely on temporal reasoning (such as QA tasks). 

We observe significant performance lift by adding VSM (L4), and the local and global alignments between subtitle and visual frames learned through VSM are especially effective on TVR and How2R. Adding additional MFFR (L5) reaches slightly worse results. Our observation is that MFFR is competing with (instead of complimentary to) MNCE during pre-training, which renders the effect of MFFR negligible. 

\paragraph{Effect of Pre-training Datasets}
We study the effect of pre-training datasets by comparing TV dataset with HowTo100M. In this study, we first pre-train our model on HowTo100M dataset (L6). We observe a performance drop on TVR, while a performance boost on TVQA, How2R and How2QA, compared to the model trained on TV dataset (L4). Our hypothesis is that text-based video-moment retrieval is more sensitive to video domains. Although HowTo100M dataset contains much more videos, the model still benefits more from being exposed to similar TV videos during pre-training.

\paragraph{Hierarchical Design vs. Flat Architecture}
To validate the effectiveness of our model design, we compare \textsc{Hero} with two baselines (with and without pre-training): ($i$) Hierarchical Transformer (\textsc{H-Trm}) baseline, constructed by simply replacing the Cross-modal Transformer with a RoBERTa model and encoding subtitles only;\footnote{The inputs to Temporal Transformer in \textsc{H-Trm} are the summation of initial frame embedding and max-pooled subtitle embeddings from RoBERTa.}
($ii$) Flat BERT-like encoder (\textsc{F-Trm}).\footnote{\textsc{F-Trm} takes as input a single sequence by concatenating the embeddings of visual frames and all subtitle sentences, and encodes them through one multi-layer Transformer.} 

For this ablation experiment, we use TVR and TVQA as evaluation tasks. Results are summarized in Table~\ref{table:model_ablation}: ($i$) Without pre-training, \textsc{F-Trm} is much worse than \textsc{Hero} on both tasks. This is due to \textsc{H-Trm} and \textsc{Hero}'s explicit exploitation of the temporal alignment between two modalities of videos. ($ii$) Pre-training lifts \textsc{Hero} performance by a large margin, but not much for \textsc{F-Trm} or \textsc{H-Trm}. This indicates that cross-modal interactions and temporal alignments learned by \textsc{Hero} through pre-training can provide better representations for downstream tasks.

\paragraph{HERO vs. SOTA with and w/o Pre-training}
We compare \textsc{Hero} with task-specifc state of the art (SOTA) models, including XML~\citep{lei2020tvr} for TVR and STAGE~\citep{lei2019tvqaplus} for TVQA. 
As shown in Table~\ref{table:model_ablation}, our model consistently outperforms SOTA models on both tasks, with or without pre-training. Note that for TVQA, STAGE is trained with additional supervision on spatial grounding with region-level features for each frame. Without additional supervisions, \textsc{Hero} is able to achieve better performance.

\begin{table}[t!]
\renewcommand\thetable{2}
\centering
\resizebox{.48\textwidth}{!}{
\begin{tabular}{l l c  c  c c }
\hline
Pre-training & Model &  & TVR& & TVQA\\ 
 \cmidrule(lr){3-5} \cmidrule(lr){6-6}  
& &  R@1  & R@10 & R@100  &  Acc.\\
\hline
\multirow{4}{*}{No\tablefootnote{\label{note:init}Model parameters are initialized with RoBERTa weights following \citet{lei2020tvr}.}} & SOTA & 2.76 &9.08 & 15.97 & 70.50\\
\cmidrule(lr){2-6}
&\textsc{F-Trm} & 1.99  & 7.76 & 13.26 &31.80 \\
&\textsc{H-Trm} & 2.97  & 10.65  & 18.68  & 70.09   \\
&  \textsc{Hero}  & 2.98 & 10.65 & 18.25 & 70.65   \\
\hline
\multirow{3}{*}{Yes} &\textsc{F-Trm}\tablefootnote{\textsc{F-Trm} is pre-trained with MLM+MNCE. VSM and FOM cannot be directly applied.} & 2.69 & 9.21 & 15.98 & 49.12\\
& \textsc{H-Trm} & 3.12& 11.08  & 18.42 &   70.03\\
 &\textsc{Hero}  & \textbf{4.44} & \textbf{14.69}  & \textbf{22.82} & \textbf{72.75}\\
\hline
\end{tabular}
}
\caption{Ablation study on model design, comparing \textsc{Hero} to a flat BERT-like encoder (\textsc{F-Trm}) baseline, a Hierarchical Transformer (\textsc{H-Trm}) baseline, and  task-specific SOTA models on TVR and TVQA val set.}
\label{table:model_ablation}
\vspace{-3mm}
\end{table}

\paragraph{Key Conclusions}

The main observations from these extensive ablation studies are summarized as follows:
\begin{itemize}
 \item The optimal pre-training setting is MLM + MNCE + FOM + VSM, when trained on HowTo100M dataset and TV dataset.
 \vspace{-2mm}
    \item FOM effectively helps downstream tasks that rely on temporal reasoning (e.g., video QA tasks). 
    \vspace{-2mm}
    \item VSM encourages frame-subtitle alignment, which is especially effective for video-moment retrieval tasks.
    \vspace{-2mm}
       \item The hierarchical design in \textsc{HERO} explicitly aligns subtitles and frames, while a flat model architecture can only learn this alignment through implicit attention.
       \vspace{-2mm}
    \item \textsc{HERO} consistently outperforms SOTA with and without pre-training, which further demonstrates the effectiveness of \textsc{HERO} model design.
\end{itemize}

\subsection{Results on Downstream Tasks}\label{sec:sota_comparison}
Table~\ref{table:results_tab} reports \textsc{Hero} results on the test splits of all downstream tasks. \textsc{Hero} is pre-trained on both TV and HowTo100M datasets, with the optimal pre-training setting: MLM + MNCE + FOM + VSM. We compare \textsc{Hero} with task-specific SOTA models on each downstream task, including: XML~\citep{lei2020tvr} for TVR, Didemo and How2R; HowTo100M~\citep{miech2019howto100m} for MSR-VTT; STAGE~\citep{lei2019tvqaplus} for TVQA; Multi-stream~\citep{liu2020violin} for VIOLIN; and MMT~\citep{lei2020tvr} for TVC. Note that we cannot directly apply STAGE to How2QA, as it was specifically designed to leverage region-level features. Our  \textsc{Hero}  model  achieves new state of the art across all benchmarks. 

\paragraph{Results on Multi-channel Tasks}
Table~\ref{tab:sota_a} shows  results on downstream tasks consisting of multi-channel videos (video + subtitle). On TVR R@1, \textsc{Hero} results nearly double those from XML.\footnote{\label{note:tvr_test}To be consistent with TVR leaderboard, results are reported on tIoU$>$0.7 without nms.}  Further, without leveraging fine-grained region-level features, \textsc{Hero} outperforms baseline models by +3.28\% on TVQA and +0.75\% on VIOLIN. When evaluated on TVC, video and subtitles are encoded by \textsc{Hero}, then fed into a 2-layer Transformer decoder to generate captions. Even though no pre-training was applied to the decoder, \textsc{Hero} surpasses SOTA baseline across all metrics, especially +4.60\% on Cider. In addition, \textsc{Hero} establishes a strong baseline for new benchmarks How2R and How2QA. 

\paragraph{Results on Single-channel Tasks}
Table~\ref{tab:sota_b} presents results on DiDeMo for text-based video-moment retrieval task and MSR-VTT for text-based video retrieval task. On DiDeMo, \textsc{Hero} surpasses XML by +0.55/+4.72/+10.65 on R@1/10/100, without leveraging Temporal Endpoint Feature used in XML. On MSRVTT, \textsc{Hero} outperforms existing video pre-training model (HowTo100M) by +1.9/+3.2/+4.9 on R@1/5/10. 

To evaluate in multi-channel setting, we also finetuned \textsc{Hero} on MSR-VTT and DiDeMo using both video channel and extracted subtitle channel (with ASR tools). When augmenting DiDeMo/MSR-VTT with ASR inputs, \textsc{Hero} performance is further improved. Although our model design focuses on “truly” multimodal videos (video+subtitle input), these results demonstrate \textsc{Hero}’s superior generalizability to different video types (multi- and single-channel). More results and analysis are provided in Appendix \ref{app:more_res}.

\section{Conclusion}
In this paper, we present a hierarchical encoder for video+language omni-representation pre-training.
Our \textsc{Hero} model presents a hierarchical architecture, consisting of Cross-modal Transformer and Temporal Transformer for multi-modal fusion.
Novel pre-training tasks are proposed to capture temporal alignment both locally and globally.
Pre-trained on two large-scale video datasets, \textsc{Hero} exceeds state of the art by a significant margin when transferred to multiple video-and-language tasks. Two new datasets on text-based video-moment retrieval and video QA are introduced to serve as additional benchmarks for downstream evaluation.
We consider extension of our model to other video-and-language tasks as future work, as well as developing more well-designed pre-training tasks.

\bibliography{emnlp2020}
\bibliographystyle{acl_natbib}

\clearpage
\appendix

\newpage
\section{Appendix}
\label{sec:appendix}

\subsection{Additional Experiments} \label{app:more_res}
\begin{table*}[t!]
\centering
\resizebox{1.0\textwidth}{!}{
\begin{tabular}{l c c  c  c c c c c c c c c c}
\hline
Method \textbackslash Task & \multicolumn{3}{c}{TVR} &   \multicolumn{3}{c}{How2R} & TVQA & How2QA & VIOLIN & \multicolumn{4}{c}{TVC} \\ 
 \cmidrule(lr){2-4} \cmidrule(lr){5-7} \cmidrule(lr){8-8} \cmidrule(lr){9-9} \cmidrule(lr){10-10} \cmidrule(lr){11-14} 
& R@1 & R@10 & R@100  &  R@1 & R@10 & R@100 & Acc. &  Acc. & Acc. &Bleu & Rouge-L & Meteor & Cider\\
\hline
SOTA baseline &2.62 & 8.45 & 14.86  & 1.97 & 8.32 & 13.45 & 70.50 & - & 67.84  & 10.53 & 32.35 & 16.61 & 44.39 \\
SOTA baseline $^{\dagger}$ &2.76 & 9.08 & 15.97  & 2.06 & 8.96 & 13.27 & -& - &  -  & 10.90 & 32.68 & 16.83 & 45.86 \\
\hline
\specialcelll{\textsc{Hero}\\w/o pre-training\footnoteref{note:init} 
} & 2.98 & 10.65 & 18.42 & 2.17 & 9.38 & 15.65 & 70.65 & 71.36 & 65.72 &10.75 & 32.72 & 16.42 & 43.62\\
\specialcelll{\textsc{Hero}\\w/ pre-training} &  \textbf{5.13} &  \textbf{16.26} & \textbf{24.55} &  \textbf{3.85} & \textbf{12.73} & \textbf{21.06}   & \textbf{74.80}  & \textbf{73.81} & \textbf{68.59} & \textbf{12.25} & \textbf{34.10} & \textbf{17.54} & \textbf{50.46}  \\
\hline
\end{tabular}
}
\vspace{-2mm}
\caption{Results on the validation set of six multi-channel video downstream tasks, compared to task-specific SOTA models: XML~\citep{lei2020tvr} for TVR and How2R, STAGE~\citep{lei2019tvqaplus} for TVQA (inapplicable to How2QA due to region-level features), Multi-stream~\citep{liu2020violin} for VIOLIN, and MMT~\citep{lei2020tvr} for TVC. $^{\dagger}$ indicates re-implementation of the model using our visual frame features.}
\label{table:results_val_tab}
\end{table*}
\begin{table*}[t!]
\centering
\resizebox{.9\textwidth}{!}{
\begin{tabular}{l c   c c c c c c c c c}
\hline
Downstream Task & Pre-training  & \multicolumn{3}{c}{Video Ret.} & \multicolumn{3}{c}{Moment Ret.\footnoteref{note:tvr_eval}} & \multicolumn{3}{c}{Video Moment Ret.\footnoteref{note:tvr_eval}}\\
 \cmidrule(lr){3-5} \cmidrule(lr){6-8} \cmidrule(lr){9-11}
& &  R@1  & R@10 & R@100  & R@1  & R@10 & R@100 & R@1  & R@10 & R@100 \\
\hline
\multirow{2}{*}{TVR} &  No\footnoteref{note:init} 

& 19.44 & 52.43 & 84.94 & 3.76& 9.59 & 61.77 & 2.98  & 10.65 & 18.25  \\
&  Yes & \textbf{30.11} & \textbf{62.69} & \textbf{87.78} & \textbf{4.02} & \textbf{10.38} & \textbf{62.93} & \textbf{5.13} &  \textbf{16.26} & \textbf{24.55}  \\
\hline
\multirow{2}{*}{How2R} &  No\footnoteref{note:init} 
& 11.15 & 39.78 & 59.62 & 4.94 & 12.73 & 67.90 &2.21 & 9.52 & 15.17   \\
&  Yes   & \textbf{14.73} & \textbf{47.69} & \textbf{68.37} & \textbf{6.48} & \textbf{15.69} & \textbf{70.38} &  \textbf{3.78} & \textbf{12.96} & \textbf{20.75}  \\
\hline
\end{tabular}
}
\caption{Detailed results on TVR and How2R val set, including the main-task (Video Moment Retrieval) and two sub-tasks (Video Retrieval and Moment Retrieval). }
\label{table:tvr_htr}
\end{table*}
For further analysis, Table~\ref{table:results_val_tab} provides comparison between \textsc{Hero} and task-specific SOTA models on the validation splits of each downstream task.\footnote{For VIOLIN, we report results on test set for fair comparison, since no validation results are reported in~\citet{liu2020violin}.} For fair comparison, we re-run XML~\citep{lei2020tvr} and MMT~\citep{lei2020tvr} experiments using our visual frame features, which achieve slightly better performance than the reported results in \citet{lei2020tvr}. Note that we cannot directly apply our frame-level visual features to STAGE~\citep{lei2019tvqaplus} and Multi-stream~\citep{liu2020violin}, which require region-level features for each video frame.

Overall, \textsc{Hero} achieves state-of-the-art results on all downstream tasks. Our model consistently outperforms XML on both TVR and How2R, with or without pre-training. Table~\ref{table:tvr_htr} also provides detailed results on TVR and How2R in three different evaluation settings from \citet{lei2020tvr}: ($i$) Video Retrieval, ($ii$) Moment Retrieval, and ($iii$) Video-moment Retrieval. For both TVR and How2R, pre-training significantly lifts model performance in all three settings.  Following \citet{chen2019uniter, lu2019vilbert}, we assess the embeddings learned in pre-training before any fine-tuning occurs. On How2R, HERO without fine-tuning achieves (2.11, 9.09, 14.83) for (R1, R10, R100). While the performance is significantly lower than the fine-tuned model (-1.62 for R1), it performs reasonably well without seeing any How2R query, indicating that HERO has learned to align videos and subtitles (pseudo-query) during pre-training.

Note that for TVQA, STAGE is trained with additional supervision on spatial grounding, which requires region-level features for each frame of the video. Without additional supervision on spatial grounding or fine-grained region-level features, \textsc{Hero} is able to achieve better performance than STAGE on TVQA dataset. We also observe that pre-training significantly boosts the performance of \textsc{Hero} across TVR, How2R and TVQA tasks. 

On How2QA, since STAGE was specifically designed to leverage region-level features, we cannot directly apply STAGE. Thus, we only compare \textsc{Hero} performance w/o and with pre-training. Results exhibit consistent patterns observed on other downstream tasks: pre-training achieves better performance than w/o pre-training. 

Pre-training greatly lifts \textsc{Hero} performance on VIOLIN by approximately $+2.9\%$. However, \textsc{Hero}, without pre-training, presents worse performance than the SOTA baseline.  Unlike Multi-stream, which leverages fine-grained region-level features, our results are reported on global frame-level features. Therefore, it may be difficult for \textsc{Hero} to capture the inconsistency between hypothesis and video content.  For example, changes of hypotheses about region-level attributes (color, shape, and etc.) may result in different conclusions.
Extending \textsc{Hero} for region-level video representations could be an interesting future direction.

\textsc{Hero} is also extensible to generation task: multi-modal video captioning. Our results on TVC show that \textsc{Hero} with pre-training surpasses MMT by a large margin. Although pre-training is only applied to the encoder, it significantly improves \textsc{Hero} performance on TVC across all metrics. When no pre-training is applied, \textsc{Hero} is slightly inferior to the SOTA baseline. Our hypothesis is that TVC has short video context (with video length of 9-second on average) but our model is designed for long video representation learning (TVR/TVQA with video length of 76-second on average). 
How to design pre-training tasks for MMT on TVC or including decoder pre-training for \textsc{Hero} are left for future works.

\subsection{Qualitative Analysis} 

\begin{figure*}[t!]
     \centering

     \begin{subfigure}[b]{\textwidth}
         \centering
         \includegraphics[height=3in]{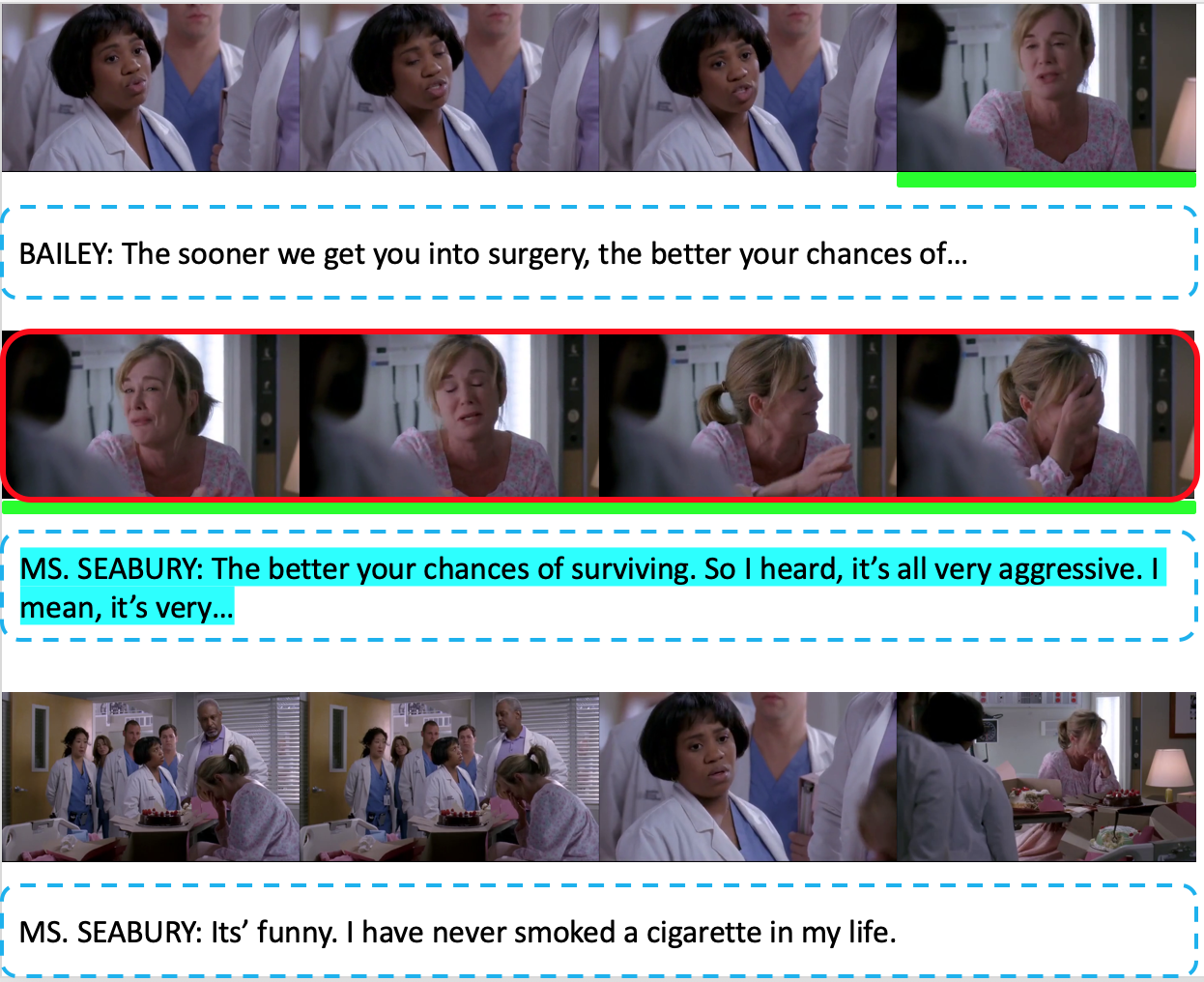}
         \caption{TV Dataset.}
         \label{subfig:vsm_tv}
     \end{subfigure}

    \begin{subfigure}[b]{\textwidth}
         \centering
         \includegraphics[height=3in]{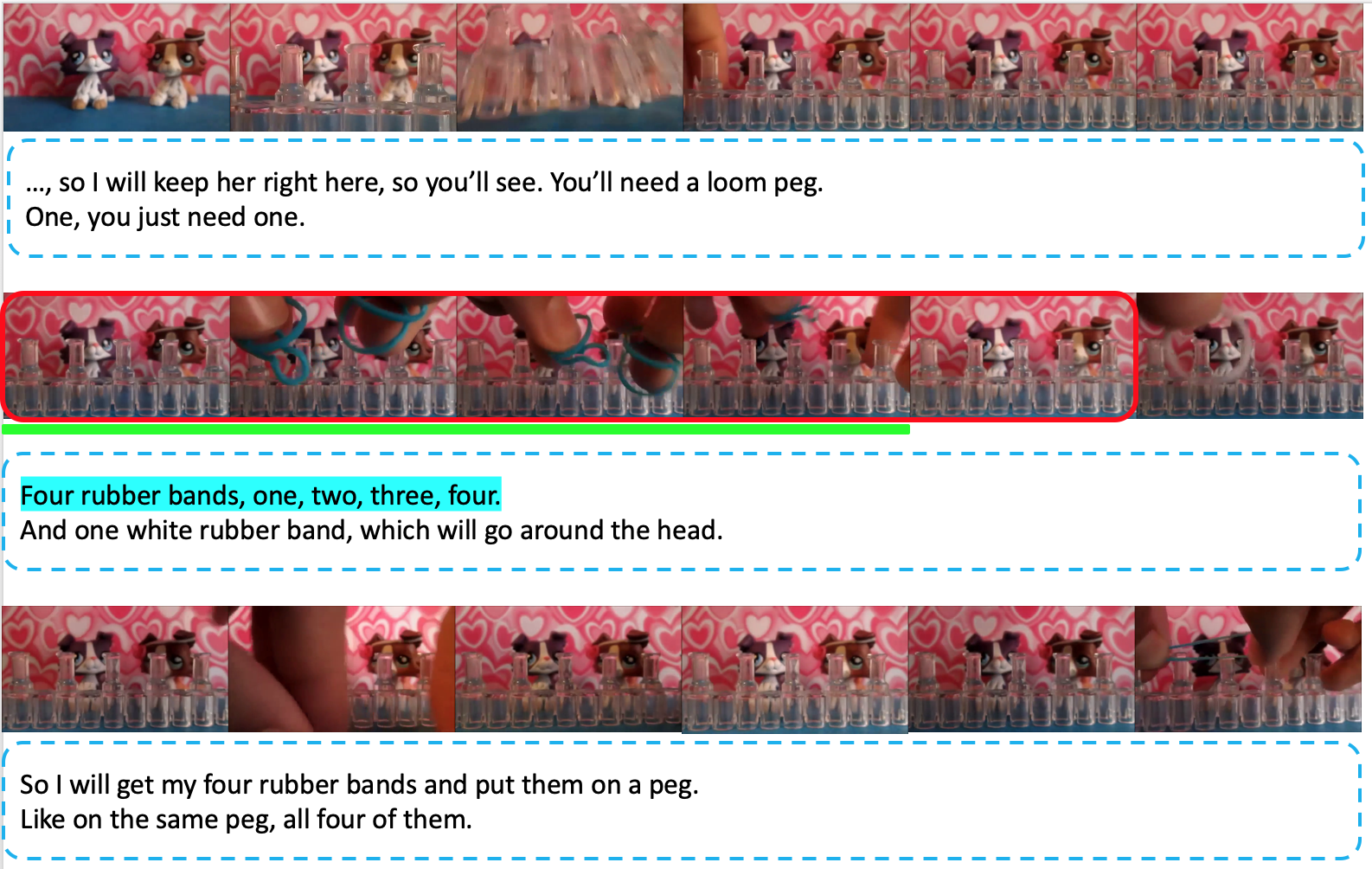}
         \caption{HowTo100M Dataset.}
         \label{subfig:vsm_ht100m}
     \end{subfigure}
\caption{Visualization of top-1 moment predictions by \textsc{Hero} model for Video-Subtitle Matching on: (a) TV Dataset; and (b) HowTo100M Dataset. Text inside the dashed boxes is the accompany subtitles, with sampled subtitle query highlighted in blue. Groundtruth is highlighted with the green bar under the video frames. Predicted moments are bounded with boxes in red. Best viewed in color.}
\label{fig:vsm_vis}
\end{figure*}

\paragraph{Visualization of VSM}
One way to understand how \textsc{Hero} aligns subtitles with video frames is to visualize the Video-Subtitle Matching pre-training task.
We provide some examples of the top-1 moment predictions for VSM on both TV and HowTo100M corpora.
As shown in Figure~\ref{fig:vsm_vis}, the predicted moments (red) largely overlap with the ground truth moments (green) with minor differences. 
In Figure~\ref{subfig:vsm_tv}, we human could probably identify the moment by the speaker information and the visual clue of character's emotion.
For Figure~\ref{subfig:vsm_ht100m}, objects (rubber bands) might be the key matching clue.
The success of \textsc{Hero} to correctly match the moments might be a positive signal that its pre-training captures those human-identified patterns, hence leads to its strong video understanding capability.
However, more thorough analysis, both quantitative and qualitative, is needed to interpret what video-language pre-trained models have learned, which we leave to future works.

\begin{figure*}[t!]
     \centering
     \begin{subfigure}[b]{0.32\textwidth}
         \centering
         \includegraphics[width=\textwidth]{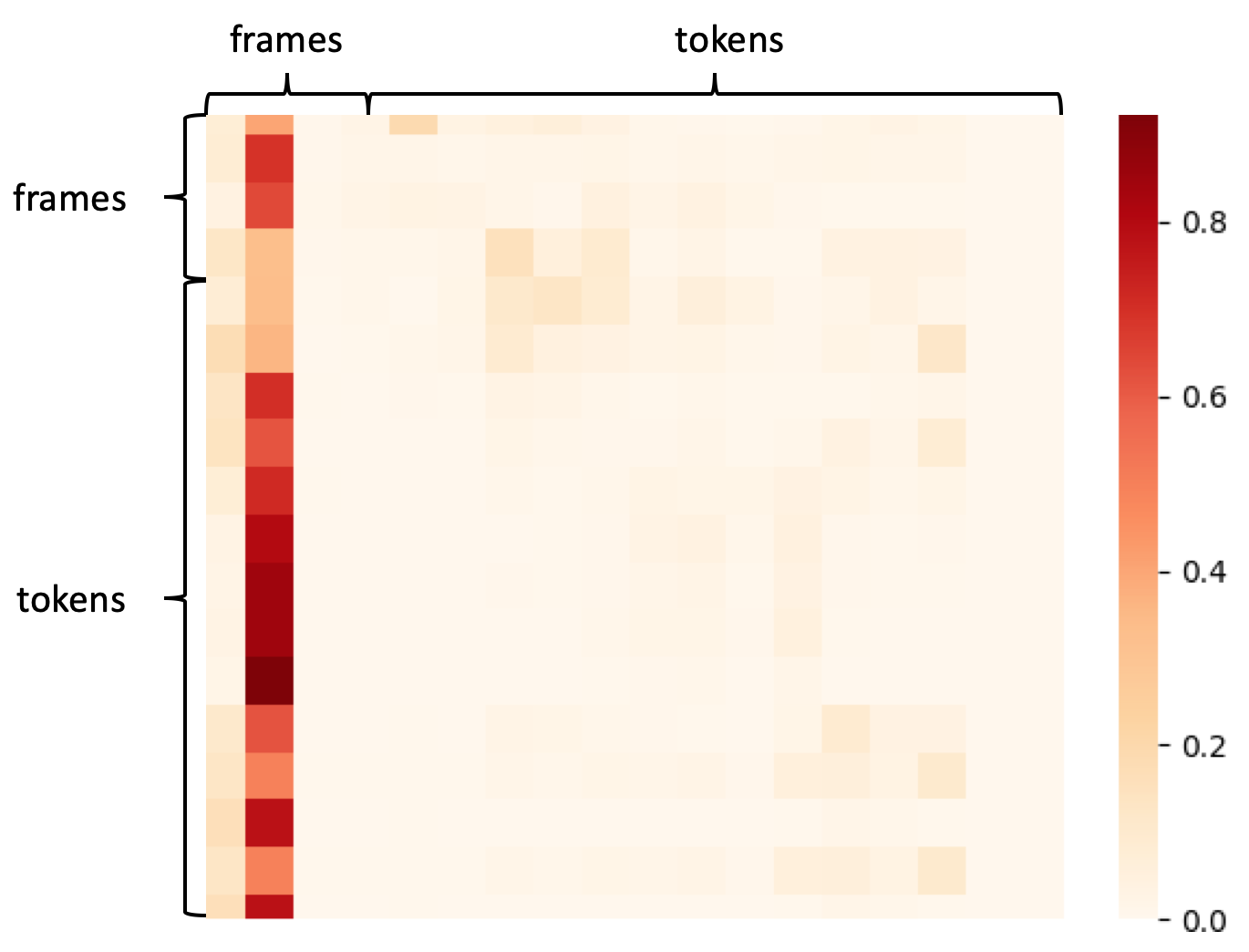}
         \caption{Vertical}
         \label{xmod:vertical}
     \end{subfigure}
     \hfill
     \begin{subfigure}[b]{0.32\textwidth}
         \centering
         \includegraphics[width=0.8\textwidth]{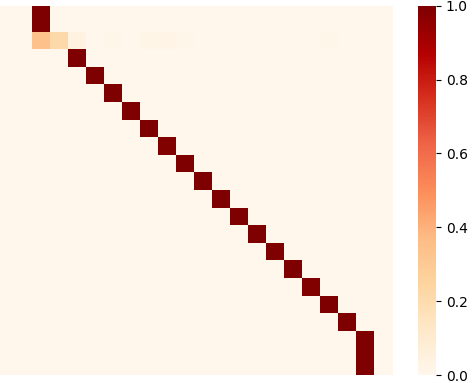}
         \caption{Diagonal}
         \label{xmod:diagonal}
     \end{subfigure}
     \hfill
     \begin{subfigure}[b]{0.32\textwidth}
         \centering
         \includegraphics[width=0.8\textwidth]{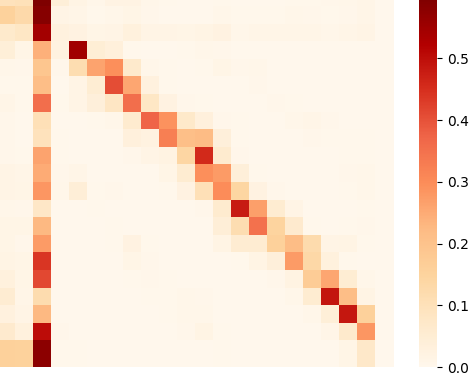}
         \caption{Vertical + Diagonal}
         \label{xmod:vnd}
     \end{subfigure}

     \begin{subfigure}[b]{0.32\textwidth}
         \centering
\includegraphics[width=0.8\textwidth]{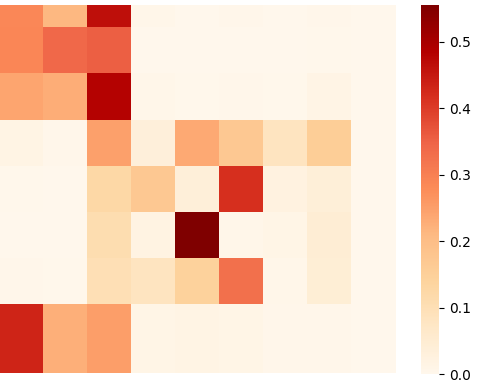}
         \caption{Block}
         \label{xmod:block}
     \end{subfigure}
     \hfill
     \begin{subfigure}[b]{0.32\textwidth}
         \centering
         \includegraphics[width=0.8\textwidth]{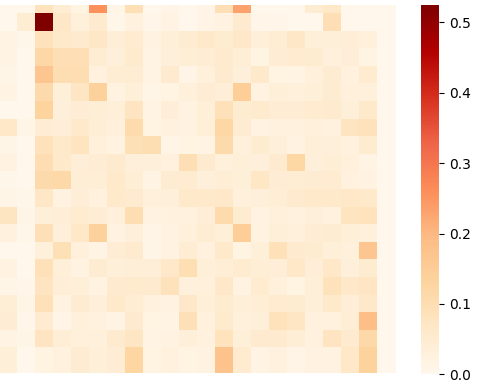}
         \caption{Heterogeneous}
         \label{xmod:heter}
     \end{subfigure}
     \hfill
     \begin{subfigure}[b]{0.32\textwidth}
         \centering
         \includegraphics[width=0.8\textwidth]{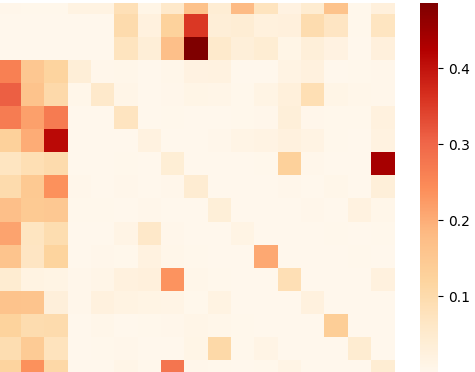}
         \caption{Reversed Block}
         \label{xmod:rblock}
     \end{subfigure}
\caption{Visualization of the attention maps learned by Cross-modal Transformers of \textsc{Hero} model.}
\label{fig:xmodal_attn}
\end{figure*}

\paragraph{Attention Pattern Visualization} Following~\citet{kovaleva2019revealing} and~\citet{chen2019uniter}, we analyze observable patterns in the attention maps of \textsc{Hero}.
Figure~\ref{fig:xmodal_attn} provides visualization examples of the attention maps learned by the Cross-modal Transformer.
For completeness, we briefly discuss each pattern here:
\begin{itemize}
  \item \textit{Vertical}: Attention to a specific frame.
  \item \textit{Diagonal}: Locally-focused attention to the token/frame itself or preceding/following tokens/frames.
  \item \textit{Vertical + Diagonal}: Mixture of Vertical and Diagonal.
  \item \textit{Block}: Intra-modality attention, \textit{i.e.}, textual self-attention or visual self-attention.
  \item \textit{Heterogeneous}: Diverse attentions that cannot be categorized and highly dependent on actual input.
  \item \textit{Reversed Block}: Cross-modality attention, \textit{i.e.}, text-to-frame and frame-to-text attention.
\end{itemize}

Note that we observe patterns slightly different from~\citet{chen2019uniter}: \textit{Vertical} patterns (Figure~\ref{xmod:vertical}) are usually over a specific frame instead of special tokens (\texttt{[CLS]} or \texttt{[SEP]}).
We leave more sophisticated attention analysis/probing to future works.

\subsection{Downstream Adaptation}
\begin{figure*}[!t]
\centering
  \includegraphics[width=0.9\linewidth]{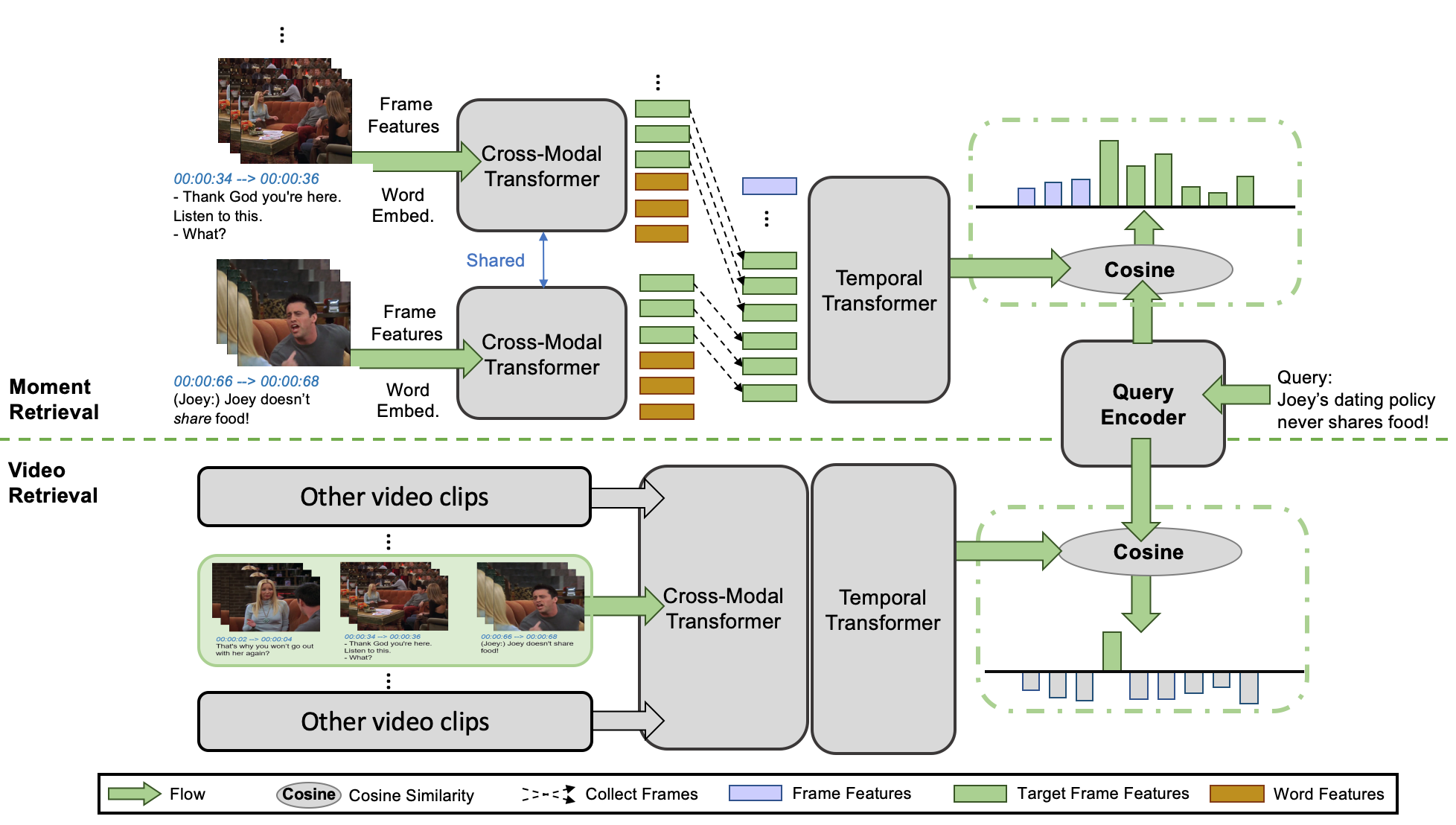}
 \caption{\textsc{Hero} model adapted to downstream task: Text-based Video Moment Retrieval.} 
  \label{fig:retrieval_model}
\end{figure*} 

\begin{figure*}[!t]
\centering
  \includegraphics[width=0.9\linewidth]{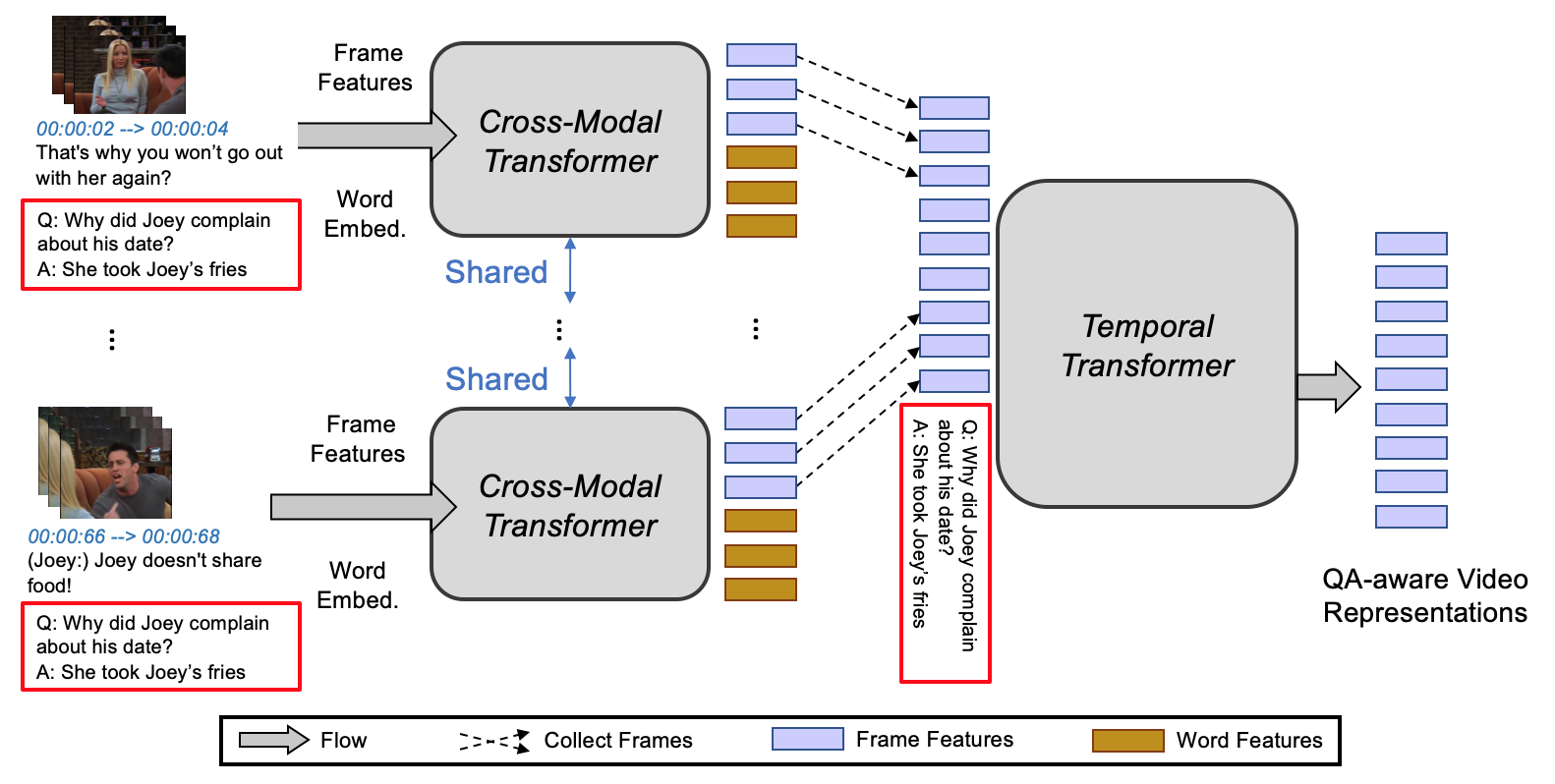}
 \caption{\textsc{Hero} model adapted to downstream task: Video Question Answering.} 
  \label{fig:qa_model}
\end{figure*} 

The pre-trained model can be readily adapted to downstream video+language tasks through end-to-end finetuning. Below, we describe the detailed adaptation approach to four downstream tasks: ($i$) text-based video moment retrieval, ($ii$) video question answering, ($iii$) video-and-language inference and ($iv$) multimodal video captioning. 

\paragraph{Text-based Video-moment Retrieval} The input video clip with accompanying subtitles is encoded by \textsc{Hero} as illustrated in Figure~\ref{fig:retrieval_model}. 
The input query is encoded by the query encoder from the VSM pre-training task.
We follow the same procedure as in VSM to compute query-video matching scores both locally (frame-level, for moment retrieval) and globally (clip-level, for video retrieval). The model is finetuned end-to-end using loss $\mathcal{L}_{\text{VSM}}$. Similarly, we let the margin $\delta = 0.1$ and set $\lambda_1 = 0.01$ and $\lambda_2 = 8$ in the loss term $\mathcal{L}_{\text{VSM}}$.

\paragraph{Video Question Answering} For Video QA, we consider the multiple-choice setting. As illustrated in Figure~\ref{fig:qa_model}, for each answer candidate, the corresponding QA pair is appended to each of the subtitle sentences and fed into the Cross-modal Transformer to perform early fusion with local textual context. In addition, these QA pairs are also appended to the input of Temporal Transformer to be fused with global video context.
We use a simple attention layer to compute the weighted-sum-across-time of the QA-aware frame representations from the Temporal Transformer output.

These final QA-aware global representations are then fed through an MLP and softmax layer to obtain the probability score $\mathbf{p}_{ans}^{(i)}$ of all the answers for question $i$. The training objective is
\begin{equation}
\mathcal{L}_{ans} = -\frac{1}{N} \sum_{i=1}^N \log \mathbf{p}_{ans}^{(i)}[y_i]\,,
\end{equation}
where $y_i$ is the index of the ground-truth answer for question $i$.
When supervision is available,\footnote{Some existing Video QA tasks require localizing `frames of interest' for the question, \emph{e.g.}, TVQA+~\citep{lei2019tvqaplus}.} we also include the span prediction loss:
\begin{equation}
\mathcal{L}_{span} = -\frac{1}{2N} \sum_{i=1}^N (\log \mathbf{p}_{st}^{(i)}[y_i^{st}] + \log \mathbf{p}_{ed}^{(i)}[y_i^{ed}])\,,
\end{equation}
where $\mathbf{p}_{st}^{(i)}$ and $\mathbf{p}_{ed}^{(i)}$ are the prediction scores of the start and end position,  obtained by applying weighted-sum-across-answers attention to the Temporal Transformer output followed by two MLPs and a softmax layer. $y_i^{st}, y_i^{ed}$ are the indices of the ground-truth start and end positions for question $i$. 

The final loss $\mathcal{L}_{\text{QA}} =  \mathcal{L}_{ans} + \lambda \mathcal{L}_{span}$, where $\lambda$ is the hyper-parameter that balance the above two terms. Empirically, we found that $\lambda = 0.5$ yields the best model performance.

\paragraph {Video-and-Language Inference} Similar to Video QA, each natural language hypothesis (or query) is appended to each of the subtitle sentences and also to the input of Temporal Transformer. A simple attention pooling layer is added to \textsc{Hero} to obtain the final query-aware global representations.

Video-and-language inference task can be regarded as a binary classification problem. We supervise the training using cross-entropy loss. 

\paragraph{Multimodal Video Captioning}
With a simple addition of a Transformer decoder~\citep{vaswani2017attention}, we can extend \textsc{Hero} for multimodal video captioning.
We feed the whole subtitle-aligned video clip into \textsc{Hero} and obtain the subtitle-fused video representation for each frame.
Next, frame representations are grouped by the ``moment of interest" using the time interval provided in the caption annotation.
The decoder-to-encoder attention is applied on the representations of the corresponding video moment and the decoder is trained with conventional left-to-right language modeling cross-entropy loss together with the \textsc{Hero} encoder end-to-end.
To make the comparison to MMT~\citep{lei2020tvr} as fair as possible, we use shallow Transformer decoder (2-layer) with 768 hidden size.
We do not use self-critical RL or its variants to optimize test metrics.
Following MMT, greedy decoding is used at inference.

\paragraph{Single-channel Tasks} 
Although \textsc{Hero} is designed for multi-channel videos (video+subtitle), we can easily extend it to single-channel video (video-only) tasks by adding an empty-string subtitle input and  pair it with the whole frame sequence. For DiDeMo, we follow the same procedure as in VSM to compute both frame-level (for moment retrieval) and  clip-level (for video retrieval) query-video matching scores. For MSR-VTT, a text-based video retrieval task, only clip-level scores are computed.

\subsection{Frames/Subtitles Pre-processing} \label{app: pairing}
Given a pair of video clip and its associated subtitle, we first extract a sequence of visual frames $\mathbf{v} = \{v_i\}^{N_v}_{i=1}$ at a fixed frame rate ($N_v$ is the number of visual frames in a video clip). The subtitle is parsed into sentences $\mathbf{s} = \{s_i\}^{N_s}_{i=1}$ ($N_s$ is the number of sentences in each subtitle). Note that $N_v \neq N_s$ in most cases, since a subtitle sentence may last for several visual frames. We then align the subtitle sentences temporally with the visual frames. Specifically, for each subtitle sentence $s_i$, we pair it with a sequence of visual frames whose timestamps overlap with the subtitle timestamp, and denote these visual frames as $\mathbf{v}_{s_i} = \{v_{s_i}^j\}_{j=1}^K$ ($K$ is the number of overlapping frames with $s_i$). In the case that multiple sentences overlap with the same visual frame, we always pair the frame with the one with maximal temporal Intersection over Union (tIoU) to avoid duplication. It is possible that a subtitle sentence is not paired with any visual frame, and in this case, we concatenate it to the neighboring sentences to avoid information loss.

\subsection{Implementation Details}
We extract 2304-dimensional Slowfast~\citep{feichtenhofer2019slowfast} features at a fixed frame rate (TV: 2/3 frame per second, HowTo100M: 1/2 frame per second). and 2048-dimensional ResNet-101~\citep{he2016deep} features at doubled frame rate and max-pooled to get a clip-level feature. The final frame features is  concatenation of the two features with dimension 4352. The model dimensions are set to (L=6, H=768, A=12) for Cross-Modal Transformer and (L=3, H=768, A=12) for Temporal Transformer, where L is the number of stacked Transformer blocks; H stands for hidden activation dimension and A is the number of attention heads. For pre-training task VSM, we let the margin $\delta = 0.1$ and set $\lambda_1 = 0.01$ and $\lambda_2 = 8$ 
in the loss term $\mathcal{L}_{\text{VSM}}$.

Our models are implemented based on PyTorch~\citep{paszke2017automatic}.\footnote{https://pytorch.org/} To speed up training, we use Nvidia Apex\footnote{https://github.com/NVIDIA/apex} for mixed precision training. Gradient accumulation~\cite{ott2018scaling} is applied to reduce multi-GPU communication overheads.
All pre-training experiments are run on Nvidia V100 GPUs (32GB VRAM; NVLink connection). We use AdamW optimizer \cite{AdamW} with a learning rate of $3e-5$ and weight decay of $0.01$ to pre-train our model. The best pre-trained model is trained on 16 V100 GPUs for about 3 weeks. Finetuning experiments are implemented on the same hardware or Titan RTX GPUs (24GB VRAM) with AdamW optimizer but different learning rates. 

\subsection{Downstream Tasks} \label{app:downstream_data}
\noindent \textbf{TVR}~\citep{lei2020tvr} is the first to introduce text-based video-moment Retrieval task for multi-channel videos (video+subtitle): given a natural language query, a model is required to not only retrieve the most relevant video clip from the video corpus, but also localize the relevant moment in the retrieved video clip. TVR is built upon the TV dataset, split into 80\% train, 10\% val, 5\% test-public and 5\% test-private. On average, 5 queries were collected for each video clip. Among them, 74.2\% of queries are related to video only, 9.1\% to text only, and 16.6\% to both video and text.

\vspace{5pt}
\noindent \textbf{TVQA}~\citep{lei2018tvqa} was introduced along with the TV dataset. Given a video clip and the accompanying subtitles, the goal is to answer a multiple-choice question about the video. Each video clip has 7 questions, with 5 answers per question. The start/end points of relevant moments are provided for each question.\footnote{\label{note:tv_split}Train, val and test video splits are the same as TVR.}

\vspace{5pt}
\noindent \textbf{VIOLIN}~\citep{liu2020violin} is a new Video-and-Language Inference task. Given a video clip with aligned subtitles as premise, a model needs to infer whether a natural language hypothesis is entailed
or contradicted by the given video clip. It consists of 95.3K video-hypothesis pairs from 15.9K video clips, split into 80\% train, 10\% val and 10\% test. 

\vspace{5pt}
\noindent \textbf{TVC}~\citep{lei2020tvr} is a multimodal Video Captioning dataset extended from TVR, containing 262K descriptions paired with 108K video moments.\footnoteref{note:tv_split}
Note that it differs from traditional video captioning tasks in that models are allowed to utilize subtitle texts as input.

\vspace{5pt}
\noindent \textbf{DiDeMo}~\citep{didemo} is designed for text-based video-moment retrieval on single-channel videos (video-only). It consists of 10.6K unedited video from Flickr with 41.2K sentences aligned to unique moments in the video. The dataset is split into 80\% train, 10\% val and 10\% test. Note that moment start and end points are aligned to five-second intervals and the maximum annotated video length is 30 seconds.

\vspace{5pt}
\noindent \textbf{MSR-VTT}~\citep{xu2016msr-vtt}, for text-based video retrieval on single-channel videos (video-only), includes YouTube videos collected from 257 popular video queries from 20 categories (e.g. music, sports, movie, etc.). It contains 200K unique video clip-caption pairs. We follow the same setup in \citet{yu2018joint} to evaluate our model on MSR-VTT.

\vspace{5pt}
\paragraph{Evaluation Metrics} Text-based Video-moment Retrieval can be decomposed into two sub-tasks: ($i$) Video Retrieval: retrieve the most relevant video clip described by the query; ($ii$) Moment Retrieval: localize the correct moment from the most relevant video clip. A model prediction is correct if: ($i$) its predicted video matches the ground-truth (in Video Retrieval); and ($ii$) its predicted span has high overlap with the ground-truth (in Moment Retrieval). Average recall at K (R@K) over all queries is used as the evaluation metric for TVR, How2R, Didemo and MSR-VTT. For TVR, How2R and Didemo, temporal Intersection over Union (tIoU) is used to measure the overlap between the predicted span and the ground-truth span.\footnote{\label{note:tvr_eval}During evaluation, the average recalls are calculated with tIoU$>$0.7. we apply non-maximal suppression (nms) with threshold 0.5 to  TVR and How2R predictions following \citet{lei2020tvr}. } 

TVQA and How2QA include 3 sub-tasks: QA on the grounded clip, question-driven moment localization, and QA on the full video clip. We only consider QA on the full video clip, as it is the most challenging setting among the three. Video clips in VIOLIN are constrained to a single, self-contained scene, hence no additional grounding annotation is provided. Accuracy is used to measure model performance on TVQA, How2QA and VIOLIN.

TVC performance is measured by standard captioning
metrics, inlcuding BLEU@4~\citep{papineni2002bleu}, METEOR~\citep{denkowski2014meteor}, ROUGE-L~\citep{linrouge}, and CIDEr-D~\citep{vedantam2015cider}.

\subsection{Vision+Language Pre-training Overview}\label{app:vl_overview}
Very recently, multimodal pre-training has gained increasing attention, especially in the image+text area.
Pioneering works such as ViLBERT~\cite{lu2019vilbert} and LXMERT~\cite{tan2019lxmert} propose to encode image and text modalities by two separate Transformers, with a third Transformer for later multimodal fusion. Compared to this two-stream architecture, 
VL-BERT~\cite{su2019vl}, Unicoder-VL~\cite{li2019unicoder}, B2T2~\cite{alberti2019fusion}, VisualBERT~\cite{li2019visualbert}, 
and UNITER~\cite{chen2019uniter} advocate single-stream architecture, where image and text signals are fused together in early stage. 
In VLP~\cite{zhou2019unified} and XGPT~\cite{xia2020xgpt}, image captioning is considered as additional downstream application, so is visual dialog in~\citet{murahari2019large}.  More recently, ViLBERT is enhanced by multi-task learning~\cite{lu201912}, Oscar~\cite{li2020oscar} enhances pre-training with image tags, and Pixel-BERT~\cite{huang2020pixel} proposes to align image pixels (instead of bottom-up features~\cite{anderson2018bottom}) with text.  Through these pre-training efforts, tremendous progress has been made for vision-and-language representation learning.

\subsection{Video+Language Tasks Overview}
Text-based Video-moment retrieval is one of the most popular video+language tasks currently studied.  \citet{anne2017localizing} and \citet{gao2017tall} introduce the task of Single Video Moment Retrieval (SVMR), which aims at retrieving a moment from a single video via a natural language query. \citet{escorcia2019temporal} extends SVMR to Video Corpus Moment Retrieval (VCMR), extending searching pool from single video to large video corpus. TVR~\citep{lei2020tvr} defines a new task, Video-Subtitle Corpus Moment Retrieval, which provides temporally aligned subtitle sentences along with the videos as inputs. For this new task, XML~\citep{lei2020tvr} is proposed to compute similarity scores between the query and each modality separately (visual frames, subtitles) and then sum them together for final prediction. 

\begin{figure}[t!]
     \begin{subfigure}[b]{0.45\textwidth}
         \includegraphics[width=\textwidth]{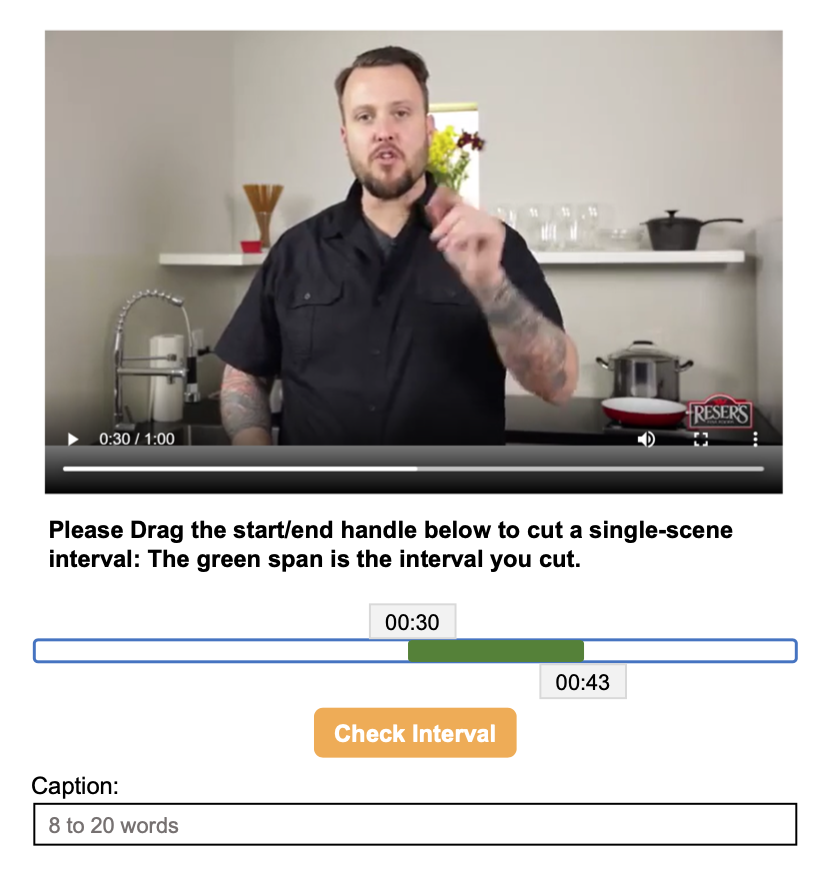}
         \caption{User interface for query annotation. Each worker is provided with a video clip and required to select a single-scene clip from the video, then write a query in the text box.}
         \label{fig:query_ui}
     \end{subfigure}
     \begin{subfigure}[b]{0.45\textwidth}
         \includegraphics[width=\textwidth]{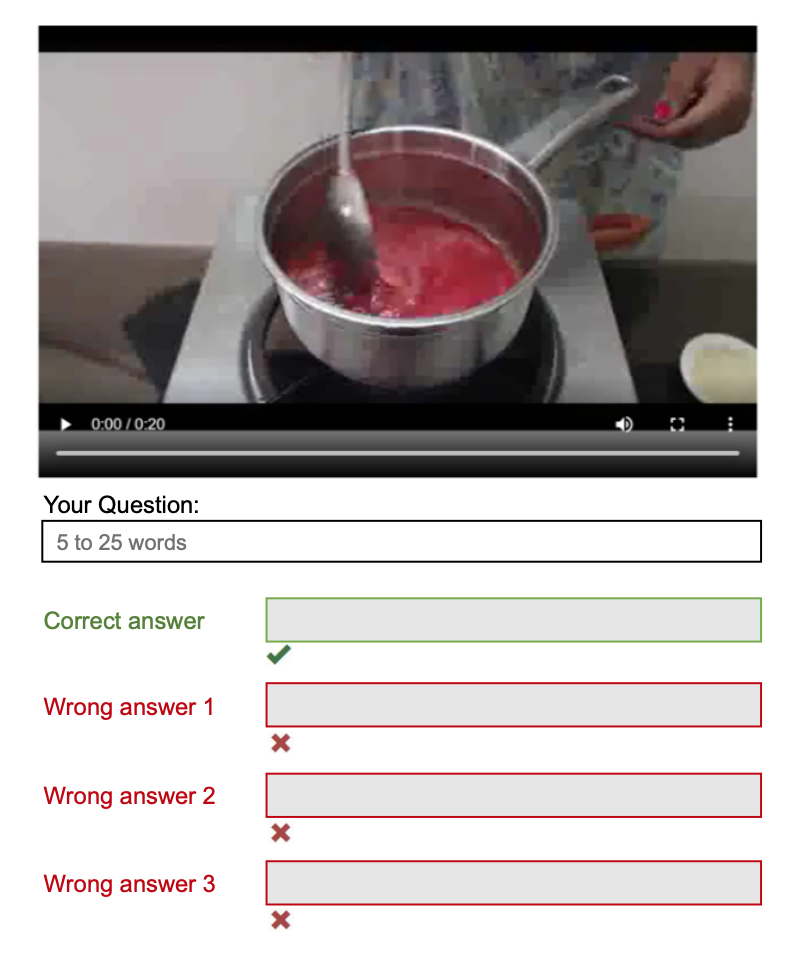}
         \caption{User interface for question/answer annotation. Each worker is provided with a segmented clip and required to write a question with four answers in the text boxes.}
         \label{fig:qa_ui}
     \end{subfigure}
\caption{\small{Data collection interface: (a) How2R; and (b) How2QA.}}
\end{figure}

Another popular task is Video Question Answering (QA), which aims to predict answers to natural language questions given a video as context. Most previous work focuses on QA pairs from one modality only. For example, MovieFIB~\citep{maharaj2017dataset} focuses on visual concepts, MovieQA~\citep{tapaswi2016movieqa} is based on text summaries, and TGIF-QA\citep{jang2017tgif} depends on predefined templates for question generation on short GIFs. TVQA~\citep{lei2018tvqa} designed a more realistic multimodal setting: collecting human-written QA pairs along with their associated video segments by providing the annotators with both video clips and accompanying subtitles. Later on, \citet{lei2019tvqaplus} augmented TVQA with frame-level bounding box annotations for spatial-temporal video QA, and introduced the STAGE framework to jointly localize moments, ground objects, and answer questions. 

Inspired by natural language inference~\citep{bowman2015large, williams2017broad} and visual entailment~\citep{xie2019visual}, \citet{liu2020violin} recently proposed Video-and-Language Inference task along with VIOLIN dataset, which requires a model to draw inference on whether a written statement entails or contradicts a given video clip. This new task is challenging to solve, as a thorough interpretation of both visual and textual clues from videos is required to achieve in-depth understanding and inference for a complex video scenario.

There are also recent studies on video captioning~\cite{venugopalan2015sequence,pan2016jointly,gan2017semantic,zhou2018end,zhou2019grounded}, popular benchmarks including Youtube2Text~\cite{guadarrama2013youtube2text}, MSR-VTT~\cite{xu2016msr}, YouCook2~\cite{zhou2018towards}, ActivityNet Captions~\cite{krishna2017dense} and VATEX~\cite{wang2019vatex}. Unlike previous work mostly focusing on captions describing the visual content, a unique TVC~\citep{lei2020tvr} dataset was released with captions that also describe dialogues/subtitles.

\begin{figure}[t!]
\centering
\includegraphics[width=0.45\textwidth]{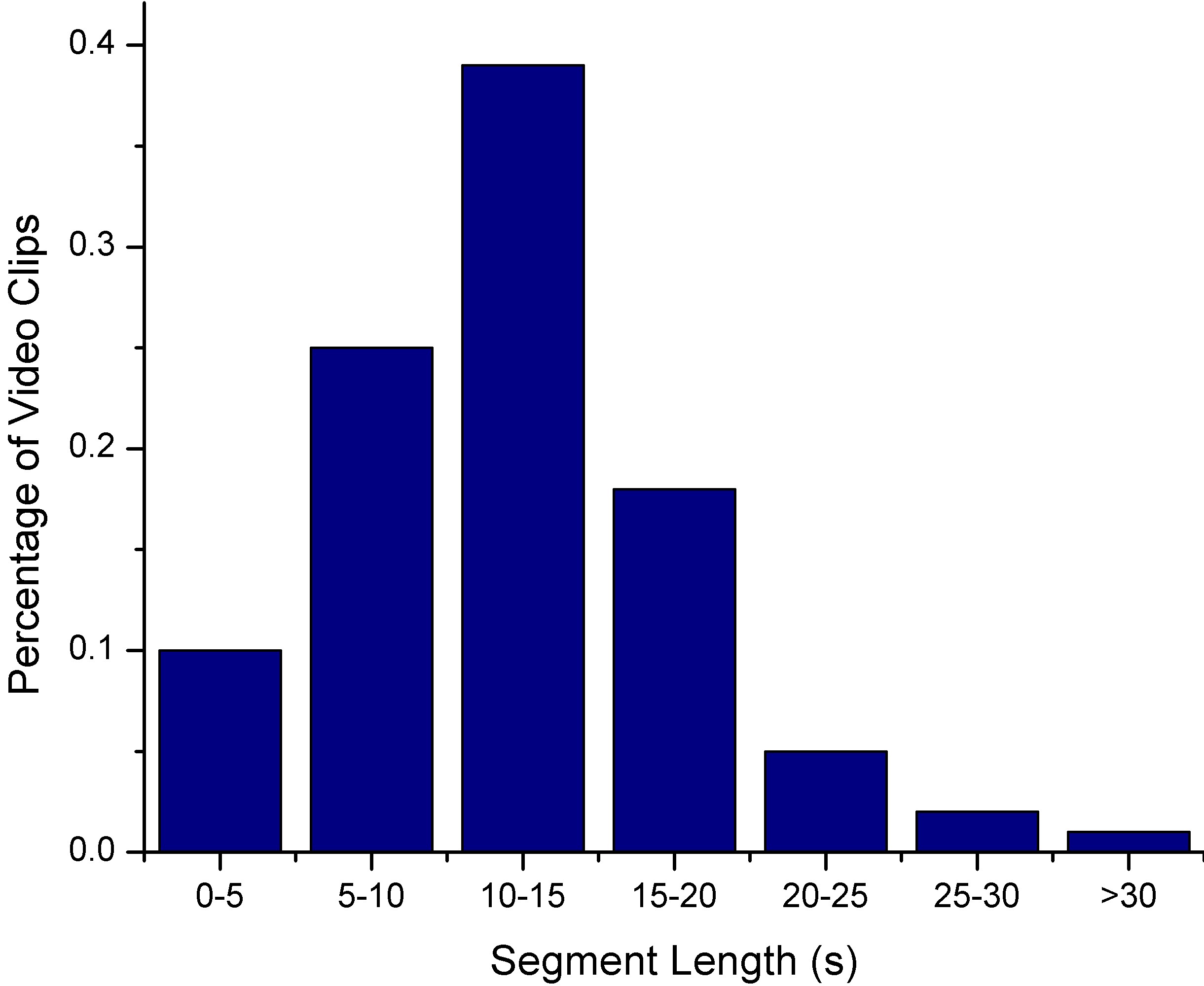}
\caption{Distribution of video segment length.}
\label{fig:video_length}
\end{figure}

\begin{figure}[t!]
\centering
\includegraphics[width=0.45\textwidth]{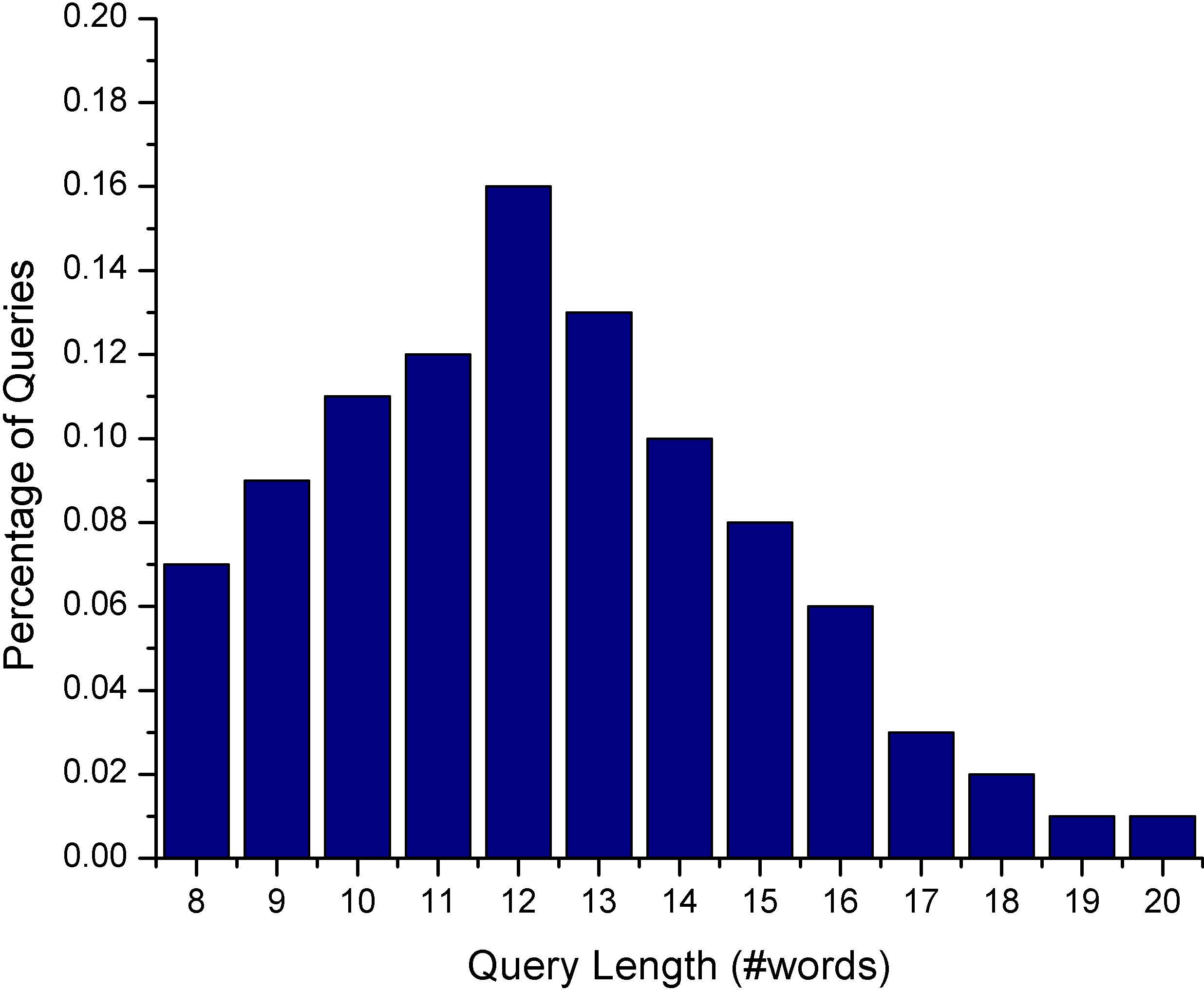}
\caption{How2R query length distribution.}
\label{fig:query_length}
\end{figure}

\begin{figure}[t!]
\centering
\includegraphics[width=0.45\textwidth]{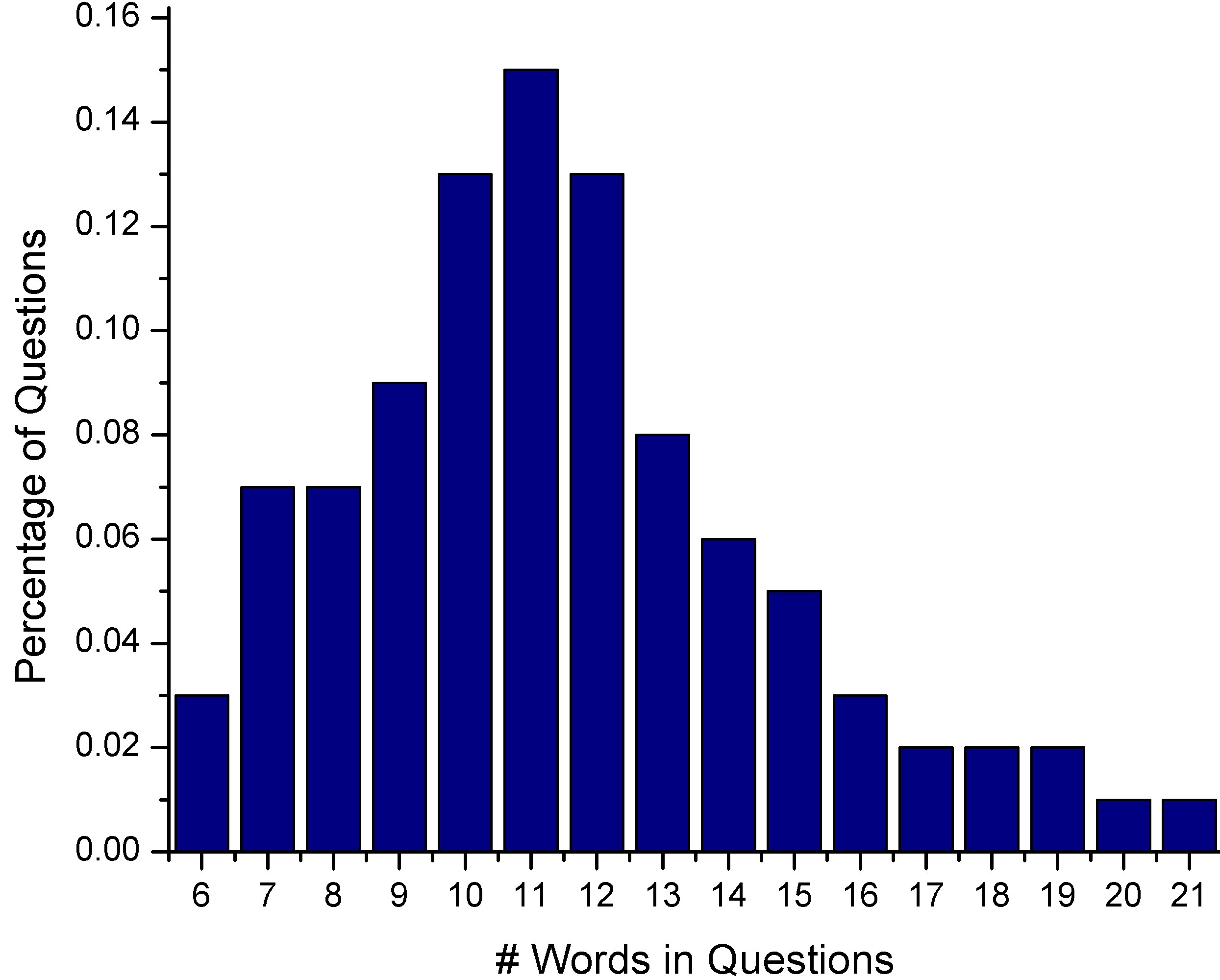}
\caption{How2QA question length distribution.}
\label{fig:question_length}
\end{figure}

\begin{figure}[t!]
\centering
\includegraphics[width=0.45\textwidth]{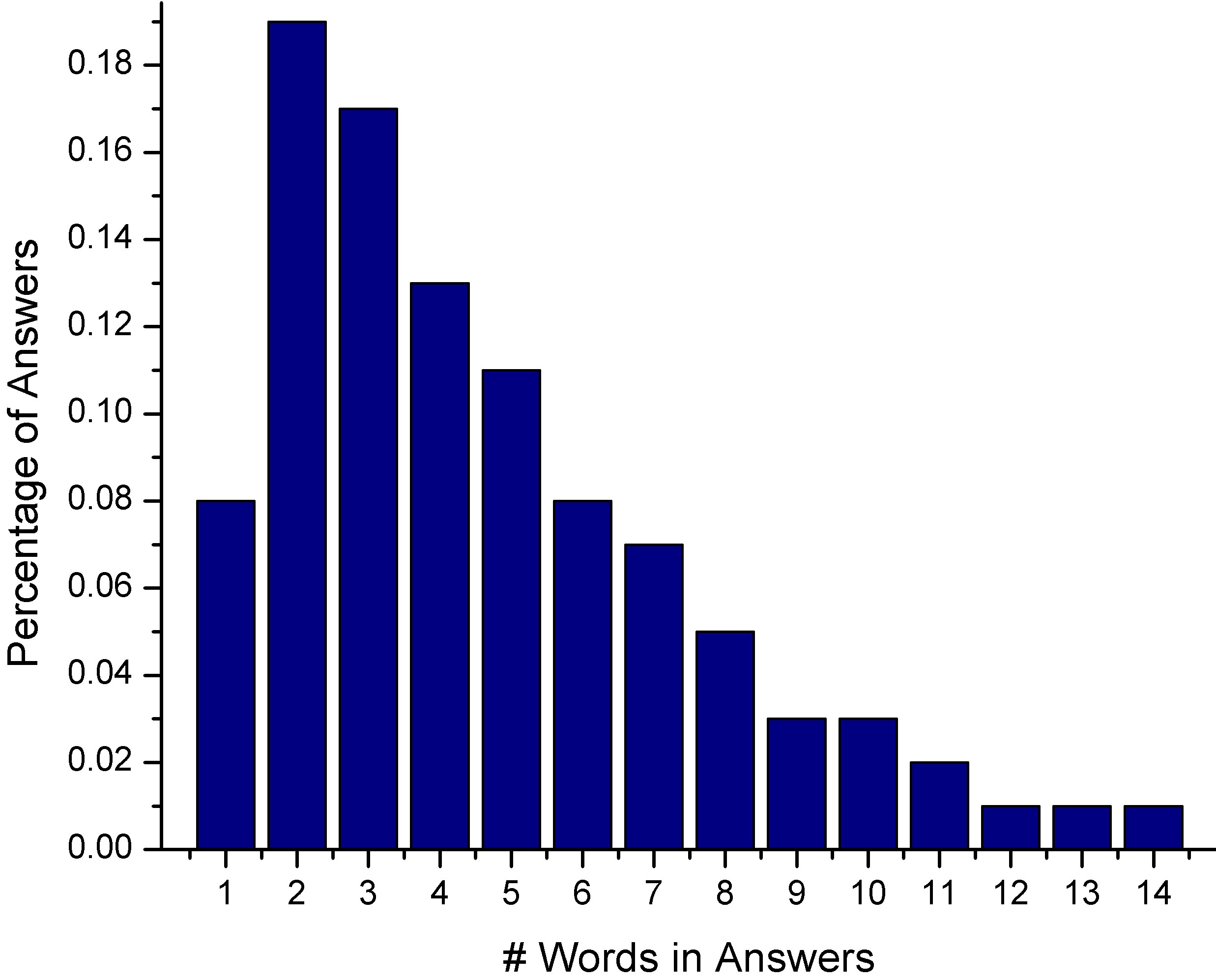}
\caption{How2QA answer length distribution.}
\label{fig:ans_length}
\end{figure}

\begin{figure}[t!]
\centering
\includegraphics[width=0.35\textwidth]{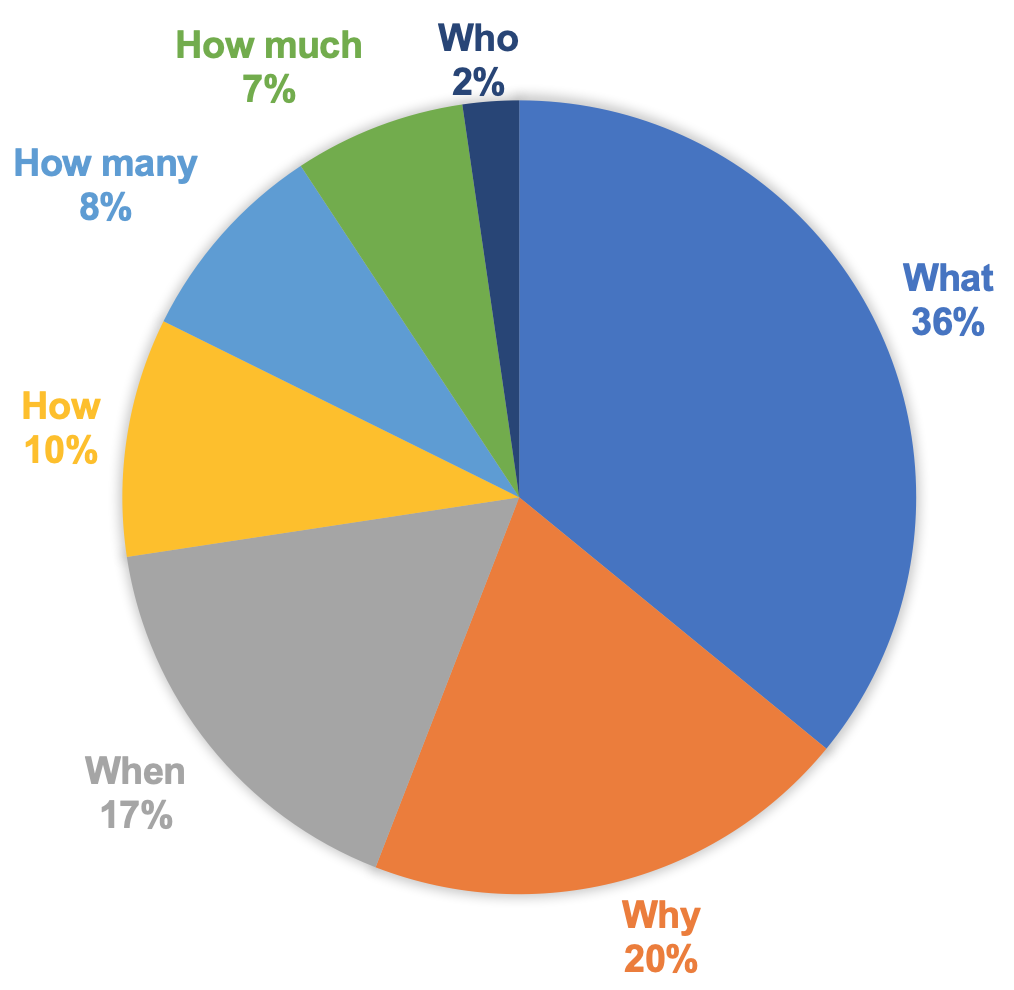}
\caption{Distribution of questions categorized by their leading words in How2QA.}
\label{fig:q_type_dist}
\end{figure}

\subsection {How2R and How2QA Benchmarks}\label{app:data_analysis}
\noindent\textbf{Data Collection Interface} Figure~\ref{fig:query_ui} and~\ref{fig:qa_ui} present the interfaces used for collecting How2R and How2QA. For How2R, the annotator is asked to first select a video segment from the presented video clip using the sliding bar, and then enter a description about the selected video segment in the text box (shown at the bottom of Figure~\ref{fig:query_ui}). For How2QA, we reuse the selected video segments collected for How2R. The annotators are asked to write a question, a correct answer and 3 wrong answers in the five text boxes shown in Figure~\ref{fig:qa_ui}.

\vspace{5pt}
\noindent\textbf{Video Segment Length Distribution} The length distribution of selected video segments is presented in Figure~\ref{fig:video_length}. The length of video segments varies from 5 to more than 30 seconds. The majority of them have length less than 15 seconds.

\vspace{5pt}
\noindent \textbf{How2R Query Length Distribution}
Figure~\ref{fig:query_length} shows the length (in number of words) distribution of collected queries in How2R. The length of queries is diverse, ranging from 8 to 20. 

\vspace{5pt}
\noindent \textbf{How2QA Question and Answer Distribution}
Figure~\ref{fig:question_length} and Figure~\ref{fig:ans_length} show the length (in number of words) distribution of collected questions and answers in How2QA. Questions are relatively longer, with more than 10 words on average. Answers are relatively shorter, most of which have less than 7 words. 

In addition, we analyze the types of collected question by plotting the distribution of their leading words in Figure~\ref{fig:q_type_dist}. In total, we collected questions in 7 different types. Majority of them starts with ``what", ``why" and ``when".

\end{document}